\documentclass[journal]{IEEEtran}
\usepackage{cite}
\usepackage{amsmath,amssymb,amsfonts}
\usepackage{algorithmic}
\usepackage{graphicx}
\usepackage{textcomp}
\usepackage{subfigure}
\usepackage{graphicx}
\usepackage{color}
\usepackage{makecell}
\usepackage{url}
\usepackage{threeparttable}
\usepackage{ulem}
\usepackage{lscape}
\usepackage{hyperref}
\usepackage{array}
\usepackage{booktabs}
\usepackage{colortbl}
\usepackage{multirow}
\usepackage{footnote}
\usepackage{threeparttable}
\usepackage{tablefootnote}
\usepackage[misc]{ifsym}
\usepackage[usestackEOL]{stackengine}
\def\BibTeX{{\rm B\kern-.05em{\sc i\kern-.025em b}\kern-.08em
    T\kern-.1667em\lower.7ex\hbox{E}\kern-.125emX}}

\usepackage{verbatim}
\usepackage{multirow}

\begin{document}

\title{NodeTrans: A Graph Transfer Learning Approach for Traffic Prediction}

\author{Xueyan Yin\textsuperscript{*}, Feifan Li\textsuperscript{*}, Yanming Shen, Heng Qi, and Baocai Yin
\thanks{Xueyan Yin, Feifan Li, and Heng Qi are with the School of Electronic Information and Electrical Engineering, Dalian University of Technology, Dalian 116024, China. }
\thanks{Yanming Shen is with the School of Electronic Information and Electrical Engineering, Dalian University of Technology, Dalian 116024, China, and also
with the Key Laboratory of Intelligent Control and Optimization for
Industrial Equipment, Ministry of Education, Dalian University of Technology,
Dalian 116024, China (e-mail: shen@dlut.edu.cn).}
\thanks{Baocai Yin is with the School of Electronic Information and Electrical
Engineering, Dalian University of Technology, Dalian 116024, China, and also
with the Peng Cheng Laboratory, Shenzhen 518055, China.}
 }

\IEEEtitleabstractindextext{%
\begin{abstract}
Recently, deep learning methods have made great progress in traffic
prediction, but their performance depends on a large amount of historical
data. In reality, we may face the data scarcity issue. In this case, deep
learning models fail to obtain satisfactory performance. Transfer learning
is a promising approach to solve the data scarcity issue. However, existing
transfer learning approaches in traffic prediction are mainly based on
regular grid data, which is not suitable for the inherent graph data in
the traffic network. Moreover, existing graph-based models can
only capture shared traffic patterns in the road network, and how to learn
node-specific patterns is also a challenge. In this paper, we propose a
novel transfer learning approach to solve the traffic prediction with few
data, which can transfer the knowledge learned from a data-rich source
domain to a data-scarce target domain. First, a spatial-temporal graph neural network is proposed, which can capture the \textit{node-specific} spatial-temporal traffic
patterns of different road networks. Then, to improve the robustness of
transfer, we design a \textit{pattern-based} transfer strategy, where we
leverage a clustering-based mechanism to distill common spatial-temporal patterns in the
source domain, and use these knowledge to further improve the prediction
performance of the target domain. Experiments on real-world datasets verify
the effectiveness of our approach.

\end{abstract}

\begin{IEEEkeywords}
Traffic Prediction, Transfer Learning, Spatial-Temporal Graph Neural Networks.
\end{IEEEkeywords}
}

\maketitle
\IEEEdisplaynontitleabstractindextext
\IEEEpeerreviewmaketitle

\section{Introduction}
\IEEEPARstart {P}{redicting} future traffic conditions (flow, speed, travel
time, etc) is one of the most urgent demands for intelligent
transportation systems. In recent years, traffic prediction has made
remarkable progress thanks to deep neural networks, which greatly improve the
ability to capture nonlinear spatial-temporal correlations. However, the
superior performance of these models relies heavily on the large amount of
data for model training, and such data requirements may sometimes not be
satisfied. For example, due to unbalanced development, some regions of a
city or even the entire city have not yet collected large amounts of
historical data. In such case, it is difficult to learn a satisfactory
prediction model using only limited historical data.

Transfer learning has recently emerged as a promising architecture for
addressing data insufficiency problem. It works by borrowing knowledge from
source domain with rich historical data to improve the performance of
target domain with scarce data, and alleviates data sparsity and cold-start problems (too few
training samples to build an initial stable model) \cite{han2020crowd}.
Transfer learning has made great progress in the field of computer vision,
but few models have been proposed in traffic prediction
\cite{2019Crowd,2019Learning,zhang2019deep}. The existing efforts for
applying transfer learning to traffic prediction have a major limitation, i.e., these methods are based
on modeling the Euclidean spatial correlations among different regions (i.e.,
grid-based data) \cite{yin2021deep}. In fact, the traffic road network structure is essentially
non-Euclidean (i.e., graph-based data) \cite{cui2020traffic}. In order to
explore the spatial information more effectively and reasonably, it is better to represent the
traffic network as graph structure. However, for grid-based transfer learning, data is sampled from the same and regular Euclidean space, and the existing solutions cannot be directly extended to graph domains with diverse graph scales and structures.

In this paper, we propose a transfer learning framework in graph domain,
\textit{NodeTrans}, to predict future traffic conditions with
scarce data. In order to capture the complex spatio-temporal correlation of traffic data in graph domain, various graph neural network models have been proposed. However, the existing methods mainly focus on capturing shared patterns of traffic road network, ignoring learning node-specific patterns that are critical for traffic prediction. In addition, due to the differences in structure and scale between different graph domains, it is difficult to find appropriate regions in the source domain that highly match the target domain for transfer. In this case, bluntly using transfer learning is not always effective and may even lead to negative effects. However, note that similar spatio-temporal patterns exist in traffic road networks, and the key is how to use spatio-temporal patterns at different nodes in the road network to transfer and design effective strategies to enhance the positive impact of transfer.

To address the above challenges, we integrate transfer learning and
spatial-temporal graph neural network into a unified model for traffic prediction. In
the spatial-temporal graph network, we first introduce the node adaptive
parameter learning process to discover \textit{node-specific} spatial-temporal traffic
patterns. Furthermore, in order to improve the
robustness of transfer, we propose a pattern-based transfer strategy, where
the common prior knowledge applicable to traffic networks can be distilled by
introducing a clustering-based mechanism into spatial-temporal graph
network. Guided by the learned cluster information, useful and reliable
knowledge can be transferred to the maximum extent. Subsequently, under the
framework of transfer learning, the learned knowledge is transferred from the
source graph domain to the target graph domain, and then only a few training
samples in the target graph domain are used for fine-tuning. Our main contributions are summarized as
follows:
\begin{itemize}
\item To the best of our knowledge, we are the first to study traffic
    prediction in the context of data scarcity in the graph domain, which
    uses transfer learning framework to learn knowledge from the data-rich
    source graph domain to facilitate prediction in the target graph domain
    with scarce data.
\item We propose a novel \textit{NodeTrans} approach which combines the
    spatial-temporal graph network and transfer learning. The former is
    used to learn \textit{node-specific} spatial-temporal traffic patterns
    in traffic road networks, while the latter adopts a
    \textit{pattern-based} strategy to enhance the positive impact of
    knowledge transfer between different domains.
\item Extensive experiments on public real-world datasets have demonstrated
    that \textit{NodeTrans} outperforms the baselines, verifying its
    effectiveness.
\end{itemize}

\section{Related Work}
\label{sec:related}

\subsection{Traffic Prediction}
Great efforts have been devoted to improve the accuracy of traffic prediction
\cite{yin2021deep}. Some earlier studies apply classical statistics and
machine learning methods, such as HA and ARIMA \cite{2003Modeling}, to
predict future traffic conditions. Recently, deep learning-based models have shown
their superior capabilities for traffic prediction. Generally, convolutional
neural network (CNN) \cite{kalchbrenner2013recurrent} and graph convolutional
network (GCN) \cite{bruna2014spectral} are used to capture the spatial
dependency of traffic data, and recurrent neural network (RNN) \cite{rumelhart1988learning}
and its variants are employed to model the nonlinear temporal dependency of traffic data. In
the latest studies, several hybrid architectures
\cite{guo2019attention,zheng2019gman,li2021spatial,ye2021coupled,Song2020Spatial,huang2020lsgcn}
combine different types of techniques to jointly model the complicated and
dynamic spatial-temporal correlations, and achieve the state-of-the-art
performance in traffic prediction.
However, the existing approaches rely on a large number of training samples, which is
not always guaranteed in practice. The problem considered in this paper is
fundamentally different, where we address the traffic prediction with scarce data samples.

\subsection{Transfer Learning}
The purpose of transfer learning is to improve the task performance of the target domain
by transferring knowledge learned from different
but related source domains \cite{han2020crowd}. 
This framework has made significant progress in the fields of computer vision and natural language processing, but it is still not well studied in traffic prediction. The method in
\cite{wei2016transfer} proposed to transfer semantically related dictionaries
learned from a data-rich source city to enrich feature representations of the
target city, and then predict air quality category of the target city. Lin et
al. \cite{lin2017transfer} designed a feature-based transfer learning model
for traffic speed prediction by extracting various hand-crafted spatial and
temporal features, and improved the interpretability of the model. However,
these two works require more empirical knowledge as guidance.

Supported by deep learning technology, several models have been proposed in recent years.
In \cite{2019Crowd}, the authors proposed an approach to transfer knowledge from a source city to
a target one by considering the latent representations of the inter-city
similar-region pairs. Zhang et al. \cite{zhang2019deep} designed a
model-based deep transfer learning framework to use the spatio-temporal
similarity of different types of cellular data to improve the prediction
accuracy. The method in \cite{2019Learning} leveraged the idea of multiple
source domains to transfer knowledge from multiple related but different
domains to obtain more comprehensive knowledge covering the target domain and
reduce the risk of negative transfer. However, these methods mainly model the
spatial correlation by decomposing the road network into grids. In fact, the
road network is essentially a graph-structured, and it is more appropriate to
represent the road network as a graph for exploring spatial information more
effectively. To the best of our knowledge, we are the first to deal with the
data-scarcity in traffic prediction based on the road network graph.

\section{Preliminaries}
\label{sec:preliminaries}
Transfer learning is the learning process of transferring knowledge from one domain (i.e., source domain) to another domain (i.e., target domain) to solve the problem of poor performance in the target domain with scare data.

In this paper, source domain refers to a traffic road network with rich historical data, which is defined as a graph $\mathcal{G}=(\mathcal{V},\mathcal{E},\mathbf{A})$, where $\mathcal{V}$ is
the set of $N$ nodes, $\mathcal{E}$ is the set of edges indicating the
connectivity between nodes, and $\mathbf{A}\in \mathbb{R}^{N\times N}$ is the
adjacency matrix. Target domain refers to a traffic road network as a graph $\mathcal{G}^{\prime}=(\mathcal{V}^{\prime},
\mathcal{E}^{\prime}, \mathbf{A}^{\prime})$  with limited historical data (e.g., 1-day),
where
$\mathcal{V}^{\prime}$ contains $N^\prime$ nodes, $\mathcal{E}^{\prime}$ is the set of edges indicating the connectivity between nodes, and $\mathbf{A}^{\prime}\in \mathbb{R}^{N^\prime\times N^\prime}$ is the adjacency matrix.

Transfer learning works as follows: (1) pre-training process, given $S$ historical traffic data on the graph $\mathcal{G}$, the goal
is to learn a function $f$ which is able to predict the traffic
conditions of the next $H$ time steps for all nodes on this graph. The
mapping relation is represented as follows:
\begin{equation}\label{1}
[\mathbf{\mathcal{X}},\mathcal{G}]\stackrel{f}{\rightarrow}\mathbf{\hat{X}}^{(t+1):(t+H)},
\end{equation}
where $\mathbf{\mathcal{X}}=\left[\mathbf{X}^{(t-S+1)},
\mathbf{X}^{(t-S+2)},\ldots,
\mathbf{X}^{(t)}\right]\in\mathbb{R}^{N\times S \times C}$,
$\mathbf{X}^{(t)}=[\mathbf{x}^{t}_{1}, \mathbf{x}^{t}_{2},\ldots, \mathbf{x}^{t}_{N}]\in\mathbb{R}^{N \times C}$ denotes the traffic condition at time step $t$, $C$ is the number of features of each node (e.g., traffic flow, speed,
etc.), and
$\mathbf{\hat{X}}^{(t+1):(t+H)}\in\mathbb{R}^{N\times H\times C}$ is the
prediction value.
(2) fine-tuning process, this step is to further optimize the
 pre-trained function $f$ learned from the source domain by using only limited data in the target domain, so as to predict the traffic condition of the next $H$ time
steps for all nodes on the graph $\mathcal{G}^{\prime}$. This process is defined as:
\begin{equation}\label{2}
[{\mathbf{\mathcal{X}}^{\prime}},\mathcal{G}^{\prime}]\stackrel{f}{\rightarrow}{\mathbf{\hat{X^\prime}}}^{(t+1):(t+H)},
\end{equation}
where ${\mathbf{\mathcal{X}^{\prime}}=\left[{\mathbf{X}^{\prime}}^{(t-S+1)},
\ldots, {\mathbf{X}^{\prime}}^{(t)}\right]}\in\mathbb{R}^{N^{\prime}\times
S\times C}$, and
${\mathbf{\hat{X^\prime}}}^{(t+1):(t+H)}\in\mathbb{R}^{N^{\prime}\times
H\times C}$ is the prediction result in the target domain.

\section{Methodology}
\label{sec:arch}
In this section, we present our graph-based transfer
learning method, $\textit{NodeTrans}$, which improves the prediction performance of the model in the target graph domain by transferring knowledge embedded in
different but related source graph domains. $\textit{NodeTrans}$ consists of a spatial-temporal graph neural network, a pattern-based transfer strategy, and a process of transferring knowledge to target domain.
\subsection{Spatial-Temporal Graph Neural Network}
\begin{figure}[t]
\centering
\includegraphics[width=1.0\columnwidth]{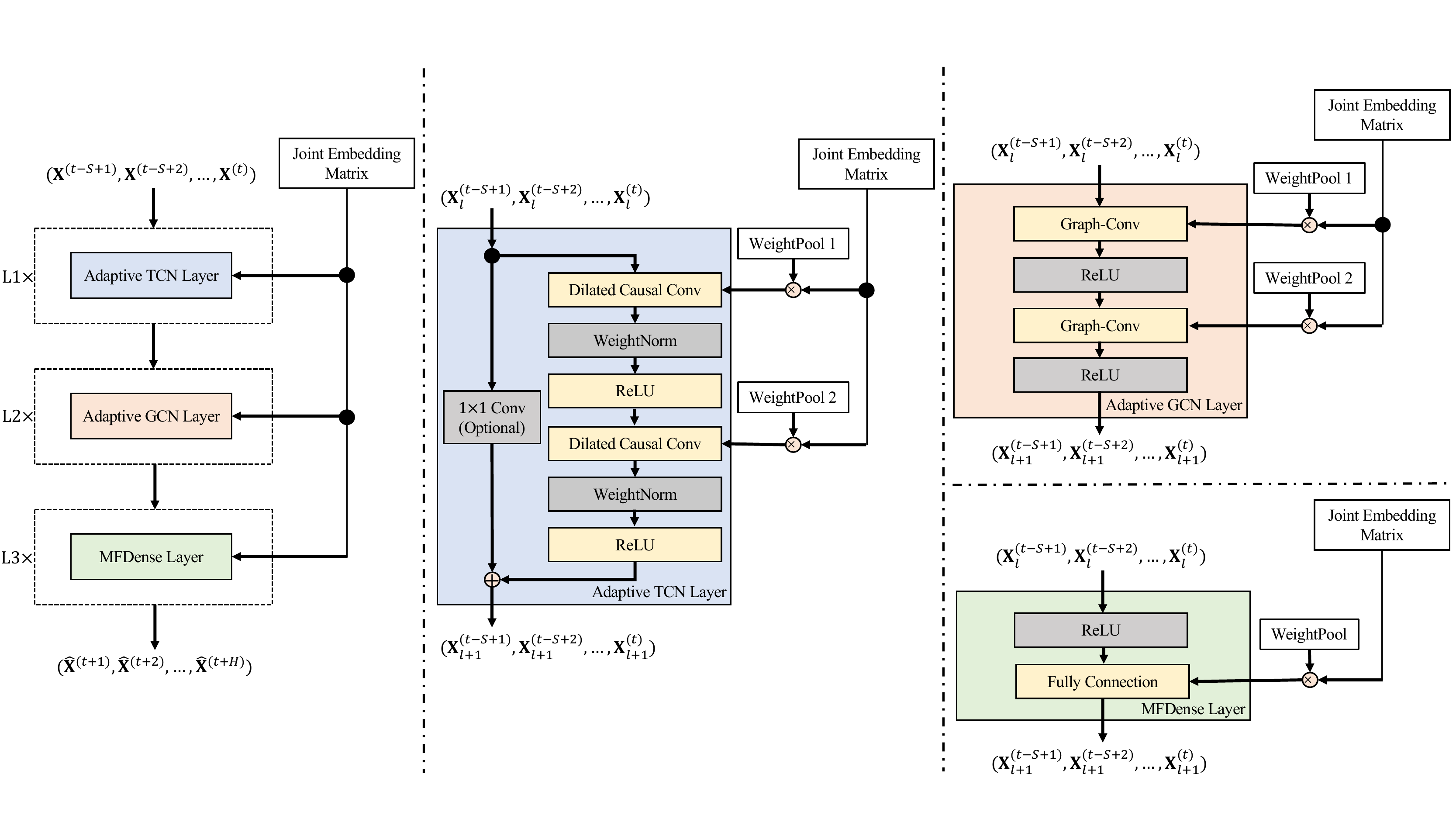}
\caption{The framework of STG-Net, where L1, L2, and L3 represent the number of stacked layers. }
\label{STG-Net}
\end{figure}

We first introduce the spatial-temporal graph network (STG-Net), which consists of adaptive TCN layer, adaptive GCN layer and MFDense layer for modeling spatio-temporal correlation and prediction, as shown in Fig. \ref{STG-Net}.

\subsubsection{Temporal Dependency Modeling}
We design an adaptive temporal convolutional networks (TCN) to model the temporal dependency of traffic data, which dynamically aggregate multiple parallel convolution kernels based on the attention weights in the implementation, instead of relying on only one single convolution kernel per layer. Assembling multiple kernels not only improves computational efficiency, but also has stronger representation power due to the fact that these kernels are aggregated in a nonlinear manner via attention \cite{chen2020dynamic}. The structure of this part consists of several stacked residual blocks, and each of which has two layers of dilated causal convolution and rectified linear unit to deal with the nonlinearity of data, and weight normalization is leveraged to the convolutional filters for normalization.  The input to this temporal modeling unit is $\mathcal{X}=\mathbf{X}^{(t-S+1):(t)}\in\mathbb{R}^{N\times S\times C}$, and the output is defined as
$\mathcal{Q}=\mathbf{Q}^{(t-S+1):(t)}\in\mathbb{R}^{N\times S\times O}$, where $O$ is the dimension of the output feature. Next, we introduce the important components in this unit.

\paragraph{Dilated causal convolution:} We leverage the dilated causal convolution \cite{yu2016multi,wu2019graph} as
our TCN to model dynamic temporal dependency.
Each layer of the network contains a 1D convolution block with an expansion
dilation factor, and different dilation factors make the model have different
receptive fields. Unlike the RNN-based method, dilated casual
convolution network can deal with long-range input sequences in a
non-recursive manner, which is beneficial to realize parallel computation and
alleviate the gradient explosion problem. Following the work of
\cite{wu2019graph}, dilated causal convolution operation is defined as:
\begin{equation}\label{5}
\mathbf{x}\ast \mathbf{f}(t)=\sum^{K-1}_{s=0}\mathbf{f}(s)\mathbf{x}(t-d_{f}\times s),
\end{equation}
where $\mathbf{x}\in \mathbb{R}^{T}$ is a 1D input sequence, $\mathbf{f}\in
\mathbb{R}^{k}$ is a filter (i.e., kernel), $K$ is the kernel length, and
$d_{f}$ is the dilation factor, which controls the skipping step length
between every two adjacent filter taps.  
\paragraph{Residual Connections} 
Considering that prediction may sometimes depend on long-range and high-dimensional historical input data, and a deeper network layer could be required. It is important to consider the stability of deeper TCNs, so we turn to the residual module to improve the performance and stability of the model. This process allows layers to learn modifications to the identity mapping rather than the entire transformation, which has proven beneficial for very deep networks, and is defined as:
\begin{equation}
	\varphi =Activation(\mathbf{x}+\mathcal{F}(\mathbf{x})),
\end{equation}
where $\mathcal{F}$ is a residual function. Note that in standard residual network \cite{zhang2017deep}, the input is added directly to the output of the residual function.
However, the input and output of TCN may have different widths, so we use a $1\times 1$ convolution to ensure that element-wise addition receives the same shape.

To further facilitate the design of transfer learning, we propose a matrix decomposition-based method to control information flow
in dilated causal convolution layers. Let
$\mathcal{X}=\mathbf{X}^{(t-S+1):(t)}\in\mathbb{R}^{N\times S\times C}$ be the input data,
and the process can be formulated as follows: 
\begin{equation}\label{14}
\mathcal{Q}=\mathbf{\Theta}\ast \mathbf{X}+\mathbf{a},
\end{equation}
where 
$\mathbf{\Theta}\in\mathbb{R}^{O\times1\times K\times C}$ and $\mathbf{a}\in\mathbb{R}$ are learnable parameters, and $\ast$
denotes the convolution operation. 
This operation transforms each node into a new output representation by a shared parameter matrix $\mathbf{\Theta}$. However, such a shared parameter matrix for all nodes is sometimes not optimal in traffic prediction \cite{bai2020adaptive}. Therefore, it is essential to learn the individual parameter space for each node, and explore its specific temporal pattern. However, if all nodes are allocated to their specific parameter space, the dimension of $\mathbf{\Theta}$ will become $\mathbf{\Theta}\in\mathbb{R}^{N\times O\times1\times K\times C}$, which is difficult to optimize due to large number of parameters and easily leads to over-fitting problem. Inspired by matrix factorization \cite{pan2019matrix,bai2020adaptive}, we decompose $\mathbf{\Theta}$ into two smaller parameter matrices (i.e., $\mathbf{E}$ and $\mathbf{U}$) instead of
learning the parameter matrix directly, where
$\mathbf{E}\in{\mathbb{R}^{N\times d}}$ is the joint embedding matrix related to
different nodes, $d$ is the embedding dimension, and $d\ll N$.
$\mathbf{U}\in{\mathbb{R}^{d\times (O\cdot 1\cdot K\cdot C)}}$ is a temporal weight pool
matrix.  Eq. (\ref{14}) can be converted to the following form:
\begin{equation}\label{25}
\mathbf{\Theta}=reshape(row\_{softmax}(\mathbf{E})\cdot\mathbf{U}),
\end{equation}
where $row\_{softmax}$ operation means to normalize the parameter matrix by row, which can narrow the range of parameters and make training eaiser.
$reshape$ operation represents the dimensional transformation of the matrix. Specifically,
assuming that $\mathbf{M}\in\mathbb{R}^{a\times (b\cdot c\cdot\\d\cdot \ldots)}$,  
$\mathbf{M}$ will become $reshape({\mathbf{M}})\in\mathbb{R}^{a\times b\times c\times d \times \ldots}$ after transformation.
Eq. (\ref{25}) can learn the specific temporal patterns of each node from a
temporal weight pool matrix.

\subsubsection{Spatial Dependency Modeling}
Graph convolution network (GCN) as an effective technique for extracting
spatial correlation is widely used in traffic prediction. Specifically, given
the input signal $\mathcal{X}=\mathbf{X}^{{(t-S+1)}:{(t)}}\in\mathbb{R}^{N\times S\times
C}$, $\mathbf{X}^{(t)}\in\mathbb{R}^{N \times C}$ and the adjacency matrix
$\mathbf{A}\in \mathbb{R}^{N\times N}$, we can obtain the output
representation $\mathcal{Z}=\mathbf{Z}^{(t-S+1):(t)}\in\mathbb{R}^{N\times S\times F}$,
the process can be defined as follows:
\begin{equation}\label{3}
\mathbf{Z}=(\mathbf{I}_{N}+\mathbf{D^{-\frac{1}{2}}AD^{-\frac{1}{2}}})\mathbf{X}\mathbf{W}+\mathbf{b},
\end{equation}
where $\mathbf{I}_{N}\in\mathbb{R}^{N\times N}$ is the identity matrix,
$\mathbf{D}\in\mathbb{R}^{N\times N}$ is the degree matrix,
$\mathbf{W}\in\mathbb{R}^{C\times F}$ and $\mathbf{b}\in\mathbb{R}^{F}$ are
the learnable parameter matrix and bias, and $F$ is the dimension of the
output feature. This GCN operation transforms each node $\mathbf{X}_{i}$ into $\mathbf{Z}_{i}$ by a shared parameter $\mathbf{W}$.
It is more reasonable to learn the individual parameter space for each node, and explore
its specific spatial pattern. However, if all nodes are allocated to their
specific parameter space, the dimension of $\mathbf{W}$ will become
$\mathbf{W}\in\mathbb{R}^{N\times F\times C}$.
Similar to Eq. \ref{25}, we
learn two smaller parameter matrices to approximately represent all the parameters in the original
$\mathbf{W}$, it can be represented as follows:
\begin{equation}\label{21}
\mathbf{W}=reshape(row\_{softmax}(\mathbf{E})\cdot \mathbf{P}),
\end{equation}
where $\mathbf{E}\in\mathbb{R}^{N\times d}$ is a joint embedding matrix the same as Eq. \ref{25}, and $\mathbf{P}\in\mathbb{R}^{d\times (F\cdot C)}$ is a spatial weight pool
matrix. Given that the pre-defined adjacency matrix usually cannot completely
cover the spatial correlation information, we use an adaptive way to learn the spatial
dependencies between each pair of nodes by multiplying $\mathbf{E}$ and
$\mathbf{E^\top}$ instead of generating $\mathbf{A}$ in advance and
calculating a Laplacican matrix. Therefore, the process can
be formulated as:
\begin{equation}
\mathbf{D}^{-\frac{1}{2}}\mathbf{A}\mathbf{D}^{-\frac{1}{2}}=softmax(ReLU(\mathbf{E}\cdot \mathbf{E}^\top)),
\end{equation}
where $softmax$ function is used to normalized the adaptive matrix. 
Here, $\mathbf{E}$ will be automatically updated to infer the spatial dependencies and obtain the adaptive matrix for graph convolution. Finally, Eq. (\ref{3}) is equivalent to the following form:
\begin{equation}\label{4}
\mathbf{Z}=(\mathbf{I}_{N}+softmax(ReLU(\mathbf{E}\cdot \mathbf{E}^\top)))\mathbf{X}\mathbf{W}+\mathbf{b},
\end{equation}
where $\mathbf{W}$ is calculated as shown in Eq. \ref{21}. This process can
be explained as each node $i$ can discover its own specific spatial pattern
from a spatial weight pool matrix $\mathbf{P}$ with the indicator of
$\mathbf{E}_{i}$. The parameter $\mathbf{P}$ contains a set of candidate
spatial patterns learned from all traffic series, whose size is actually
independent of the road network size.

\subsubsection{Prediction Layer}
Existing models usually use a predictor with shared parameters to predict the traffic conditions of all nodes in the road network \cite{pan2019matrix}. However, due to the specifics of different nodes, it is essential to learn a predictor with a separate set of parameters for each node, respectively.
Similar to the previous parts, the matrix factorization based dense layers (MFDense Layer) are designed for prediction, and a predictor with an individual parameter set is designed for each node.  Specifically, we use a weight matrix
$\mathbf{V}=[\mathbf{V}_{1},\mathbf{V}_{2},\ldots,\mathbf{V}_{N}]\in\mathbb{R}^{N\times
H\times S}$, $\mathbf{V}_{i}\in\mathbb{R}^{H\times S}$ as a transformation matrix, and $\mathbf{V}$ can be decomposed into $\mathbf{E}$ and $\mathbf{R}$, which is expressed as:
\begin{equation}
\mathbf{V}=reshape(row\_{softmax(\mathbf{E})\cdot \mathbf{R}}),
\end{equation}
where $\mathbf{R}\in\mathbb{R}^{d\times (H\cdot S)}$ is a predictor weight
pool matrix, $\mathbf{E}$ is the joint embedding matrix the same as Eq. \ref{21} and Eq. \ref{25},
By combining the node embedding matrix $\mathbf{E}$ into the temporal modeling unit, spatial modeling unit and prediction layer, we can effectively obtain the node specific spatio-temporal pattern and predictor parameters. Finally, the future traffic conditions can be predicted by the
following:
\begin{equation}\label{7}
\mathbf{\hat{X}}=\mathbf{V}\ast \mathbf{Z}+\mathbf{c},
\end{equation}
where $\mathbf{\hat{X}}$ is the prediction results, $\mathbf{c}$ is
node-specific bias. The loss function used for multi-step prediction is defined
as :
\begin{equation}\label{8}
\mathcal{L}_{p}(\mathcal{Y},\mathcal{\hat{Y}}; \mathbf{\Theta}_{total})=
\frac{\sum\limits_{i=1}^{N}\sum\limits_{j=1}^{H}\sum\limits_{k=1}^{C}||\mathbf{\hat{X}}^{(t+j)}_{ik}-\mathbf{X}^{(t+j)}_{ik}||^{2}}{N\times H\times C},
\end{equation}
where $\mathcal{Y}=\mathbf{X}^{(t+1):(t+H)}$,
$\mathcal{\hat{Y}}=\mathbf{\hat{X}}^{(t+1):(t+H)}$, and $\mathbf{\Theta}_{total}$ is
all learned parameters in the source domain.

\subsection{Pattern-based Transfer Strategy}
Directly using transfer learning is not always positive, because the
transferred knowledge may sometimes not promote the target task. Just
like in a traffic prediction scenario, we cannot directly transfer the
knowledge learned from one city to another city, because of the differences in data
distribution between different cities. If similar source
domain knowledge is leveraged in the target domain, the prediction will be
facilitated, otherwise it may be harmful. In this case, it is crucial to improve the knowledge matching between source domain and target domain to enhance the robustness of transferred knowledge. For this reason, we take into
account the different spatio-temporal patterns existing in the road network,
and propose a novel pattern-based transfer strategy to improve the
effect of transfer.

In traffic prediction, for different regions at different time steps, the spatio-temporal dependency between two nodes is different, and therefore the influence of the neighborhood on the predicted node is also different. However, there are similar spatio-temporal patterns involved in these changes. For example, school areas usually have high traffic volume when students go to or leave school, and hospital are with higher traffic volume on weekday mornings than afternoons.
Such \textit{semantic information} can be regarded
as a global property shared by different domains.
Therefore, no matter how diverse the road network structure is, the
spatio-temporal patterns between different domains will be similar. However,
in the target domain, it is difficult to capture these spatio-temporal
patterns with limited data.

To address this issue, we leverage a clustering-based mechanism in the
source domain to encourage different spatio-temporal pattern representations
to form clusters. The joint embedding matrix $\mathbf{E}$ contains specific parameters of each node,
and these parameters reflect different spatio-temporal dependencies of each
node in the road network.  Since many nodes have the same or similar spatio-temporal
patterns, in order to learn the compact representations to facilitate
generalization and transfer in the target domain, we cluster
these spatio-temporal representations of all the nodes in source domain to
$G$ categories, where the well-known K-means algorithm is used in our implementation.
\begin{figure}
\centering
\includegraphics[width=1.0\columnwidth]{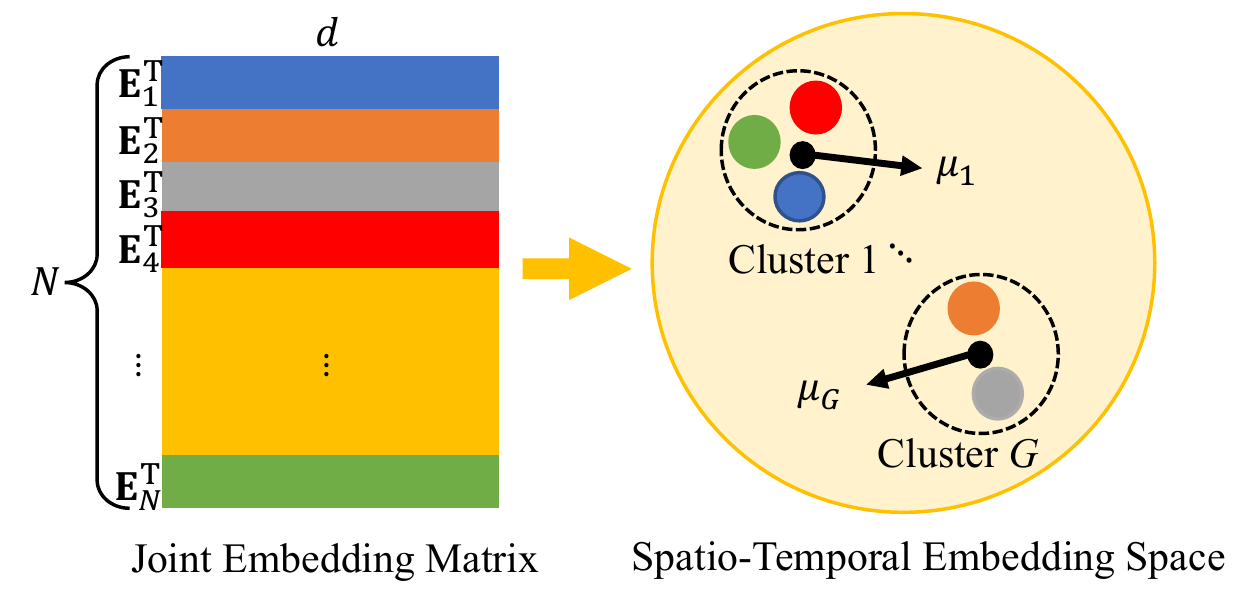}
\caption{Spatio-temporal pattern clustering.}
\label{st}
\end{figure}
As shown in Fig. \ref{st}, the left side is the joint embedding matrix $\mathbf{E}$, each row represents a node in the road network, which is clustered into $G$
different categories according to the spatio-temporal pattern of each node, as shown on the right.
Then, the cluster center learned from the source domain is initialized in the target domain.
Finally, a clustering-based regularization term is used to capture different spatio-temporal patterns
in the target domain:
\begin{equation}\label{9}
\mathcal{R}(\mathbf{E}^{\prime},\mathbf{\mu})=\frac{1}{N^{\prime}d}\sum_{i=1}^{N^{\prime}}||\mathbf{e}^{\prime}_{i}-\mathbf{\mu}_{z_{i}}||^{2},
\end{equation}
where
$\mathbf{E}^{\prime}=[\mathbf{e}^{\prime}_{1},\mathbf{e}^{\prime}_{2},\ldots,\mathbf{e}^{\prime}_{N^\prime}]^\top$ represents the joint embedding matrix of the target domain,
$\mathbf{e}^{\prime}_{i}\in\mathbb{R}^{d}$, $\mu=[\mu_{1},\mu_{2},\ldots,\mu_{G}]$, $\mu_{z_{i}}\in\mathbb{R}^{d}$, $z_{i}\in\{1,2,\ldots,G\}$ is a discrete latent variable indicating the cluster (i.e.,
spatio-temporal pattern) to which the node belongs. Thus, the cluster center
information is also included in the transferred parameters. Under the
guidance of the spatio-temporal cluster centers learned from the source domain,
the opportunity for useful and reliable knowledge transfer can be maximized,
and the possibility of negative transfer to the target domain can be reduced. In addition, it is not required that the categories of the source domain and the target domain are consistent. In fact, the spatio-temporal pattern categories that are most similar to the source domain are found from the target domain in a data-driven manner.

\subsection{Transfer Knowledge to Target Domain}
\begin{figure}[t]
\centering
\includegraphics[width=1.0\columnwidth]{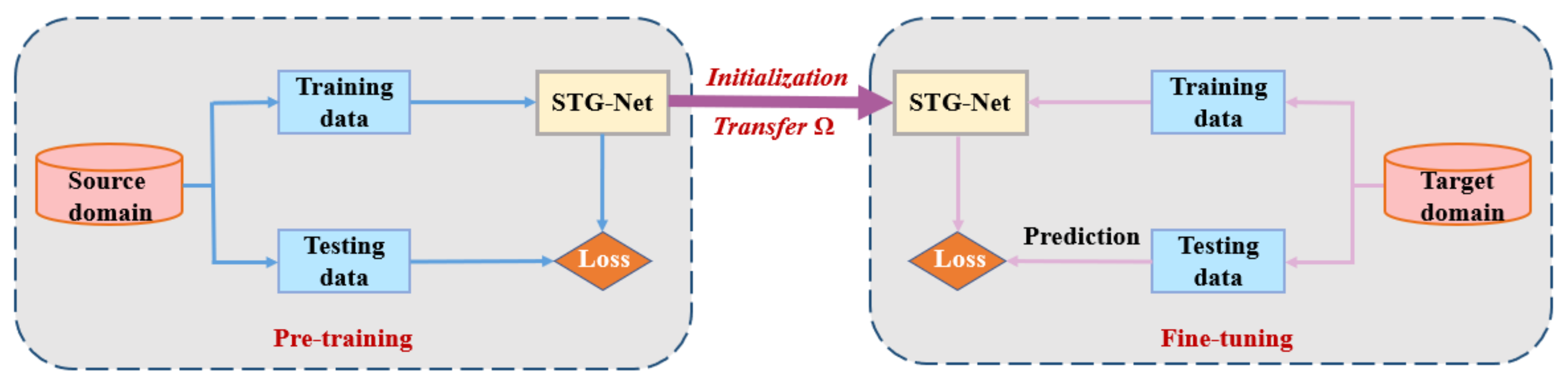} 
\caption{Transfer learning framework.}
\label{Transfer}
\end{figure}

Fig. \ref{Transfer} shows the workflow of the transfer learning framework. With
pre-training, we can first learn a set of network parameters in a source
domain with sufficient historical data. Since the number of nodes in
different road networks is different, we only transfer parameters in the
source domain that are independent of the number of nodes. For simplify, we
collectively refer these parameters as $\mathbf{\Omega}$, and then
$\mathbf{\Omega}$ is directly applied to the target domain as the
initialization parameter. Afterwards, we fine-tune the parameter $\mathbf{E}^{\prime}$
related to the number of network nodes, and update the parameters of
$\mathbf{U}$, $\mathbf{P}$, $\mathbf{R}$, $\mu$, with limited small
amount of data (e.g., only a few days) in the target domain, instead of
re-training all the parameters. Finally, the prediction performance of our
model needs to be verified on the testing data of the target domain, and the
overall loss function can be defined as:
\begin{equation}\label{10}
\mathcal{L}(\mathbf{\Pi})=\mathcal{L}_{p}(\mathcal{Y^{\prime}}, \mathcal{\hat{Y}^{\prime}};
\mathbf{\Theta}^{\prime})+\alpha \cdot \mathcal{R}(\mathbf{E}^{\prime},\mathbf{\mu}),
\end{equation}
where the cluster center estimate at iteration $t$ is obtained via $\mu^{t}=\beta\hat{\mu}+(1-\beta)\mu^{t-1}$, $\beta$ is a smoothing weight, $\hat{\mu}$ is the cluster center of current state.
 $\mathbf{\Pi}$ is all learned parameters in the target domain, $\alpha$
is the trade-off parameter, $\mathcal{R}$ is defined in Eq. (\ref{9}), and
\begin{equation}\label{11}
\begin{split}
\mathcal{L}_{p}(\mathcal{Y^{\prime}}, \mathcal{\hat{Y}^{\prime}};
\mathbf{\Theta}^{\prime})=\frac{\sum\limits_{i=1}^{N^{\prime}}\sum\limits_{j=1}^{H}\sum\limits_{k=1}^{C}||\mathbf{\hat{X^{\prime}}}^{(t+j)}_{ik}-\mathbf{{X}^{\prime}}^{(t+j)}_{ik}||^{2}}{N^{\prime}\times
H\times C},
\end{split}
\end{equation}
where $\mathcal{Y}^{\prime}=\mathbf{X^{\prime}}^{(t+1):(t+H)}$ defines the ground truth in the target domain,
$\mathcal{\hat{Y}^{\prime}}=\mathbf{\hat{X^{\prime}}}^{(t+1):(t+H)}$ represents the predicted value, and
$\mathbf{\Theta}^{\prime}$ is the parameter set learned by STG-Net in the
target domain. Based on the limited data in the target domain, our model finds
approximate spatio-temporal patterns in the source domain, and adjusts the
loss function of Eq. (\ref{10}) to improve the accuracy of pattern matching.
Note that our model can be directly extended to the multi source domains
scenario, where node specific spatio-temporal patterns under different
source domains can be learned and transferred to the target domain.

\section{Experiments}
\label{sec:exp} We evaluate the performance of \textit{NodeTrans} on four
highway datasets, including PEMSD4, PEMSD8, PEMS-BAY and METR-LA, and two
urban datasets, including JiNan and XiAn in China. Among those highway datasets, the former three datasets are
collected from Caltrans Performance Measure System (PeMS)
\cite{chen2001freeway}, and the METR-LA dataset is collected from loop
detectors of Los Angeles County \cite{li2018dcrnn}. Both urban datasets are
collected by Didi Chuxing GAIA Initiative (https://gaia.didichuxing.com). We
adopt the same data pre-processing as in \cite{guo2019attention},
\cite{li2018dcrnn} and \cite{Guo2021hierarchical}. For each source domain,
the datasets of PEMSD4, PEMSD8, JiNan and XiAn are divided into training,
validation and testing in a 6:2:2 ratio, and the datasets of PEMS-BAY and METR-LA are split into 70\% for training, 10\% for validation,
20\% for testing. For each target domain, we
respectively use 1-day, 3-day and 7-day data for training, 1-day data for
validation, and the remaining 20\% data for testing. Each sensor represents a node in a
road network graph. The sensors in PEMSD4 and PEMSD8 contain traffic flow
historical observations, while other sensors record traffic speed
observations. All highway datasets are aggregated into 5-minutes interval,
and urban datasets are sampled at 10-minutes interval. The details are shown in Table \ref{dataset}.

\begin{table*}
\centering
\caption{Details of the datasets.}
\begin{tabular}{l|p{1.4cm}p{1.4cm}|p{1.4cm}p{1.4cm}|p{1.4cm}p{1.4cm}}
\hline
\hline
Dataset&PEMSD4&PEMSD8&PEMS-BAY&METR-LA&JiNan&XiAn\\
\hline
\#Nodes&307&170&325&207&561&792\\
\#Edges&340&295&2369&1515&4797&6883\\
\#Time steps&16992&17856&52116&34772&52286&52286\\
Data type&Flow&Flow&Speed&Speed&Speed&Speed\\
Time span&1/1/2018-28/2/2018&1/7/2016-31/8/2016&1/1/2017-31/5/2017&1/3/2012-30/6/2012&1/1/2018-31/12/2018&1/1/2018-31/12/2018\\
\cline{2-7}
Time interval&\multicolumn{4}{c|}{5 min}&\multicolumn{2}{c}{10 min}\\
\cline{2-7}
Daily range&\multicolumn{6}{c}{0:00-24:00}\\
\hline
\hline
\end{tabular}
\label{dataset}
\end{table*}
\subsection{Baselines}
We compare \textit{NodeTrans} with the following widely used classical time
series models and deep learning models. Considering that there is no traffic prediction work for transfer in graph domain, and that the underlying assumption of existing models requires a large amount of training data, so we use sufficient data to train baselines in the target domain, and the data scale is the same as \textit{NodeTrans} pre-training in the source domain.

\begin{itemize}
\item HA: Historical Average models the traffic condition as a seasonal
    process, and uses the average of observed weekly patterns for
    prediction.
\item ARIMA \cite{2003Modeling}: Auto-Regressive Integrated Moving Average
    is a powerful time series analysis method for predicting future values.
\item LSTM \cite{hochreiter1997long}: Long Short-Term Memory Network is a variant of RNN method to learn sequence data. 
    \item STGCN \cite{yu2018spatio}: Spatial-Temporal Graph Convolution Network
    integrates graph convolution and 1D convolution together to extract spatial and temporal features.
\item Graph WaveNet \cite{wu2019graph}: Graph WaveNet combines graph
    convolution with dilated casual convolution to capture spatial-temporal
    dependencies.
\item AGCRN \cite{bai2020adaptive}: Adaptive Graph Convolutional Recurrent
    Network automatically captures node-specific spatial and temporal
    dynamics in traffic data.
\item HGCN \cite{Guo2021hierarchical}: Hierarchical Graph Convolution
    Network simultaneously considers the micro layers of road networks and the
    macro layers of region networks.
\item STG-Net: This is an important component of our \textit{NodeTrans}
    framework, which combines adaptive graph convolutional networks and
    dilated causal convolution networks to extract spatial and temporal
    features.
\end{itemize}
\subsection{Settings}
For hyper-parameters setting, we utilize grid search to find the optimal set
of hyper-parameters. From $d$=\{2,3,4,\ldots,15\}, $G$=\{1,2,3,4,5,10,15\}, $\alpha$=\{0,0.1,0.2,0.5,1.0,1.5\}, batch size=\{16,32,64,128\}, we record
the optimal ones in our experiment. Specifically, the number of clusters for
spatio-temporal patterns is set to 5, and the batch size is set to 64. The
embedding dimension of $d$ is set to 10, the trade-off parameter $\alpha$
is set to 1.0, and smoothing weight parameter $\beta$ is set 0.2. We stack two TCN residual blocks with
dilation factors of 1, 2 respectively. The kernel length $K$ in TCN is 3. 
Adam optimizer is used to train the
model with an initial learning rate of 0.003 and a decay rate of 0.3. The maximum values of training
epoch of pre-training and fine-tune are 200 and 400 respectively.

We adopt widely used evaluation metrics, including Root
Mean Squared Error (RMSE), Mean Absolute Error (MAE), and Mean Absolute
Percentage Error (MAPE), to verify the performance of all models.

\begin{table*}
\caption{Results of traffic speed prediction between METR-LA and PEMS-BAY.}
\label{Comparison3}
\centering
\setlength{\tabcolsep}{0.4mm}{
\begin{tabular}{l|p{1.5cm}p{1.5cm}p{1.5cm}|p{1.5cm}p{1.5cm}p{1.5cm}}
	\hline
    \hline
	\multirow{1}{*}{Method}  &  \multicolumn{3}{c|}{PEMS-BAY}&\multicolumn{3}{c}{METR-LA}\\
\cline{2-7}
\textbf{Target Data Only}&RMSE&MAE&MAPE&RMSE&MAE&MAPE\\
							\hline
HA	    &7.93&3.52&8.19\%&14.37&7.37&18.56\%\\
ARIMA	&6.50&3.38&8.30\%&13.23&6.90&17.40\%\\
LSTM	&4.96&2.37&5.70\%&8.69&4.37&13.20\%\\
STGCN	&5.69&2.49&5.79\%&9.40&4.59&12.70\%\\
Graph WaveNet &4.52&1.95&4.63\%&7.37&3.53&10.01\%\\
AGCRN   &4.54&1.96&4.64\%&7.51&3.62&10.38\%\\
HGCN &4.77&2.12&5.01\%&7.84&3.91&11.11\%\\
STG-Net(our)&4.69&2.10&4.91\%&7.68&3.74&10.82\%\\
\hline
&\multicolumn{3}{c|}{METR-LA $\rightarrow$ PEMS-BAY}&\multicolumn{3}{c}{PEMS-BAY $\rightarrow$ METR-LA}\\
\cline{2-7}
\textbf{Source \& Target Data}&RMSE&MAE&MAPE&RMSE&MAE&MAPE\\
\hline
\textit{NodeTrans} (1-day)&6.79&3.17&7.83\%&11.96&5.87&16.92\%\\
\textit{NodeTrans} (3-day)&5.82&2.69&6.71\%&10.51&5.43&15.54\%\\
\textit{NodeTrans} (7-day)&5.80&2.63&6.47\%&9.30&4.83&14.18\%\\
\hline
\hline
	\end{tabular}}
	\end{table*}

\begin{table*}
\caption{Results of traffic speed prediction between JiNan and XiAn.}
\label{Comparison4}
\centering
\setlength{\tabcolsep}{0.4mm}{
\begin{tabular}{l|p{1.5cm}p{1.5cm}p{1.5cm}|p{1.5cm}p{1.5cm}p{1.5cm}}
	\hline
    \hline
	\multirow{1}{*}{Method}  &  \multicolumn{3}{c|}{XiAn}&\multicolumn{3}{c}{JiNan}\\
\cline{2-7}
\textbf{Target Data Only}&RMSE&MAE&MAPE&RMSE&MAE&MAPE\\
							\hline
HA	    &8.94&5.93&20.84\%&8.73&5.72&20.56\%\\
ARIMA	&7.24&5.04&18.26\%&7.01&5.09&18.24\%\\
LSTM	&6.53&4.52&17.42\%&6.26&4.30&17.38\%\\
STGCN	&5.52&3.73&14.56\%&5.44&3.71&15.03\%\\
Graph WaveNet &5.10&3.44&13.22\%&5.16&3.52&13.77\%\\
AGCRN   &8.52&4.35&13.75\%&8.84&4.80&14.73\%\\
HGCN &4.85&3.24&12.52\%&5.02&3.36&13.31\%\\
STG-Net(our)&5.13&3.47&13.44\%&5.44&3.79&14.85\%\\
\hline
&\multicolumn{3}{c|}{JiNan $\rightarrow$ XiAn}&\multicolumn{3}{c}{XiAn $\rightarrow$ JiNan}\\
\cline{2-7}
\textbf{Source \& Target Data}&RMSE&MAE&MAPE&RMSE&MAE&MAPE\\
\hline
\textit{NodeTrans} (1-day)&7.03&4.93&18.20\%&6.92&4.86&18.34\%\\
\textit{NodeTrans} (3-day)&6.43&4.37&16.73\%&6.10&4.22&16.28\%\\
\textit{NodeTrans} (7-day)&6.06&4.08&16.44\%&5.82&4.01&15.75\%\\
\hline
\hline
	\end{tabular}}
	\end{table*}

\begin{table*}
\caption{Results of traffic flow prediction between PEMSD4 and PEMSD8.}
\label{Comparison5}
\centering
\begin{threeparttable}
\setlength{\tabcolsep}{0.4mm}{
\begin{tabular}{l|p{1.5cm}p{1.5cm}p{1.5cm}|p{1.5cm}p{1.5cm}p{1.5cm}}
	\hline
    \hline
	\multirow{1}{*}{Method}  &  \multicolumn{3}{c|}{PEMSD8}&\multicolumn{3}{c}{PEMSD4}\\
\cline{2-7}
\textbf{Target Data Only}&RMSE&MAE&MAPE&RMSE&MAE&MAPE\\
							\hline
HA	    &52.04&34.86&24.07\%&59.24&38.03&27.88\%\\
ARIMA	&43.23&24.02&15.89\%&58.05&35.19&20.68\%\\
LSTM	&34.06&22.20&14.20\%&41.59&27.14&18.20\%\\
STGCN	&27.09&17.50&11.29\%&34.89&21.16&13.83\%\\
Graph WaveNet &27.87&17.54&10.47\%&37.09&23.15&13.52\%\\
AGCRN   &25.22&15.95&10.09\%&32.26&19.83&12.97\%\\
HGCN \tnote{1}&--&--&--&--&--&--\\
STG-Net(our)&26.92&17.31&11.41\%&32.22&20.46&14.25\%\\
\hline
&\multicolumn{3}{c|}{PEMSD4 $\rightarrow$ PEMSD8}&\multicolumn{3}{c}{PEMSD8 $\rightarrow$ PEMSD4}\\
\cline{2-7}
\textbf{Source \& Target Data}&RMSE&MAE&MAPE&RMSE&MAE&MAPE\\
\hline
\textit{NodeTrans} (1-day)&32.18&21.50&20.22\%&41.83&27.02&23.60\%\\
\textit{NodeTrans} (3-day)&30.66&19.62&14.80\%&36.34&23.24&17.13\%\\
\textit{NodeTrans} (7-day)&28.54&18.35&12.48\%&35.22&22.31&16.08\%\\
\hline
\hline
	\end{tabular}}
	\begin{tablenotes}
	\footnotesize
	\item[1] Note that 
since the HGCN method itself requires geo-location information, but PEMSD4 and PEMSD8 datasets do not contain relevant information, they are not applicable to this method, so “--” is used instead.
       \end{tablenotes}
	\end{threeparttable}
	\end{table*}

\begin{table*}
\caption{Evaluation results between METR-LA and PEMS-BAY.}
\label{ablation_high}
\centering
\setlength{\tabcolsep}{0.4mm}{
\begin{tabular}{l|lll|lll|lll|lll|lll|lll}
	\hline
    \hline
	\multirow{2}{*}{Method}  &  \multicolumn{9}{c|}{METR-LA $\rightarrow$ PEMS-BAY}&\multicolumn{9}{c}{PEMS-BAY $\rightarrow$ METR-LA}\\
\cline{2-19}
 &\multicolumn{3}{c|}{1-day}  & \multicolumn{3}{c|}{3-day}&\multicolumn{3}{c|}{7-day} &\multicolumn{3}{c|}{1-day}&\multicolumn{3}{c|}{3-day}&\multicolumn{3}{c}{7-day}\\
    \cline{2-19}
&RMSE&MAE&MAPE&RMSE&MAE&MAPE&RMSE&MAE&MAPE&RMSE&MAE&MAPE&RMSE&MAE&MAPE&RMSE&MAE&MAPE\\
							\hline
\textbf{Target only}&&&&&&&&&&&&&&&&&&\\
LSTM&11.00&6.02&14.72\%&8.13&4.07&9.34\%&7.85&3.99&9.13\%&13.05&8.24&20.75\%&10.35&5.79&17.09\%&10.68&6.11&16.12\%\\
STGCN&8.87&4.79&11.97\%&7.14&3.83&9.05\%&6.91&3.51&8.68\%&11.53&6.51&19.14\%&10.75&6.03&17.75\%&10.21&5.77&15.33\%\\
Graph WaveNet&7.89&3.77&8.98\%&6.46&2.84&7.23\%&6.35&2.80&7.12\%&12.22&6.11&17.73\%&11.13&5.75&16.17\%&9.72&4.95&14.72\%\\
AGCRN&7.32&3.66&8.79\%&6.43&3.17&7.78\%&6.26&2.93&7.23\%&12.15&6.06&17.45\%&11.21&5.76&16.15\%&9.65&4.93&14.43\%\\
HGCN&12.20&9.93&20.61\%&8.25&5.57&12.52\%&5.99&2.84&7.06\%&12.10&6.69&19.56\%&11.33&6.44&18.61\%&9.69&4.98&15.65\%\\
STG-Net(our) &8.39&3.89&10.67\%&6.78&3.06&7.67\%&6.12&2.72&7.02\%&13.06&6.21&17.84\%&11.52&5.88&16.20\%&9.54&4.96&15.67\%\\
\textbf{Source \& Target Data}&&&&&&&&&&&&&&&&&&\\
\textit{NodeTrans}-NC&7.11&3.42&8.66\%&6.08&2.79&7.16\%&5.96&2.70&6.88\%&12.02&6.04&17.67\%&11.03&5.66&16.12\%&9.42&4.90&14.32\%\\
\textit{NodeTrans}&6.79&3.17&7.83\%&5.82&2.69&6.71\%&5.80&2.63&6.47\%&11.96&5.87&16.92\%&10.51&5.43&15.54\%&9.30&4.83&14.18\%\\
\hline
\hline
	\end{tabular}}
	\end{table*}

\subsection{Experimental Results}
Table \ref{Comparison3} and Table \ref{Comparison4} show the performance
comparison of traffic speed prediction on highway and urban datasets with
different models over the future 12 time steps ($H=12$), and Table \ref{Comparison5} shows the
results of traffic flow prediction on highway datasets in next hour. We can have the following observations: First, the deep learning model
is usually superior to the classical time series method due to its stronger
nonlinear learning ability, and our proposed model can achieve a good performance even with a small amount of training data in the
target domain. Second, since the LSTM model does not consider the spatial
correlation of the road network, its performance is slightly inferior to
other deep learning models which capture both spatial and temporal
dependencies in most of the time. Third, we see that STG-Net is superior to
or comparable to other spatio-temporal graph neural networks on all datasets,
which proves the effectiveness of our model in modeling spatial and temporal
correlations. Finally, we see that although $\textit{NodeTrans}$ is not trained using sufficient data on the target road network, its accuracy is better than or comparable to other traffic prediction methods that use sufficient data training in the target domain. Moreover,
as the amount of data in the target domain increases, the performance
of our model improves rapidly and achieves competitive results. 

\subsubsection{Ablation Study}
The clustering mechanism is the core of pattern-based
transfer strategy. To further investigate the effectiveness of transfer
strategy, we compare $\textit{NodeTrans}$ with its variants
$\textit{NodeTrans}$-NC on two different prediction tasks, where $\textit{NodeTrans}$-NC means without clustering mechanism. The results are
also shown in Table \ref{ablation_high} and
Table \ref{ablation_high1}. By incorporating the clustering
mechanism, we can see that the performance of $\textit{NodeTrans}$ has
been greatly improved, which demonstrates its effectiveness. In particular, the smaller the amount of training data in the target
domain, the greater the improvement of the model. For example, $\textit{NodeTrans}$ improves performance on RMSE by about 4.5\% when PEMS-BAY is used as the target domain dataset and 1-day data is used for fine-tuning.
Furthermore, the two tables also show that the performance of different deep learning methods when there is only a small amount of data training on the target domain. The results are not ideal, even if $\textit{NodeTrans}$-NC does not include a clustering mechanism, its performance is superior to those of deep learning models trained with only a small amount of data, further illustrating the superiority and necessity of transfer learning.
In addition, we observe that the
performance of STG-Net is much lower than that of
\textit{NodeTrans-NC}, which proves the validity of the transferred
parameters selected for the transfer process.

\begin{table*}
\caption{Evaluation results between PEMSD4 and PEMSD8.}
\label{ablation_high1}
\centering
\setlength{\tabcolsep}{0.4mm}{
\begin{tabular}{l|lll|lll|lll|lll|lll|lll}
	\hline
    \hline
	\multirow{2}{*}{Method}  &  \multicolumn{9}{c|}{PEMSD4 $\rightarrow$ PEMSD8}&\multicolumn{9}{c}{PEMSD8 $\rightarrow$ PEMSD4}\\
\cline{2-19}
 &\multicolumn{3}{c|}{1-day}  & \multicolumn{3}{c|}{3-day}&\multicolumn{3}{c|}{7-day} &\multicolumn{3}{c|}{1-day}&\multicolumn{3}{c|}{3-day}&\multicolumn{3}{c}{7-day}\\
    \cline{2-19}
&RMSE&MAE&MAPE&RMSE&MAE&MAPE&RMSE&MAE&MAPE&RMSE&MAE&MAPE&RMSE&MAE&MAPE&RMSE&MAE&MAPE\\
							\hline
\textbf{Target only}&&&&&&&&&&&&&&&&&&\\
LSTM&63.84&44.61&32.51\%&59.54&39.64&24.34\%&53.59&33.15&19.10\%&88.31&59.91&43.64\%&48.98&31.72&21.46\%&42.90&27.44&19.56\%\\
STGCN&45.44&31.78&25.34\%&38.86&26.07&17.70\%&36.42&23.93&17.10\%&47.88&32.09&27.04\%&41.21&27.27&22.41\%&39.36&26.00&20.00\%\\
Graph WaveNet&42.93&23.09&14.14\%&33.26&21.40&13.78\%&32.46&21.08&13.74\%&44.99&28.32&24.18\%&40.34&26.48&21.51\%&40.24&26.17&20.18\%\\
AGCRN&34.80&23.41&16.92\%&33.00&22.08&15.59\%&32.78&21.25&12.59\%&46.52&31.18&24.24\%&37.72&24.10&17.84\%&36.77&22.99&16.59\%\\
HGCN&--&--&--&--&--&--&--&--&--&--&--&--&--&--&--&--&--&--\\
STG-Net(our) &42.39&29.54&28.30\%&35.48&24.24&19.79\%&29.99&19.51&13.88\%&50.32&35.49&36.22\%&41.81&28.28&22.65\%&35.67&22.72&16.55\%\\
\textbf{Source \& Target Data}&&&&&&&&&&&&&&&&&&\\
\textit{NodeTrans}-NC&33.62&22.88&21.29\%&31.87&20.96&15.80\%&29.89&19.41&13.58\%&42.31&27.31&23.40\%&36.95&23.52&17.33\%&35.54&22.61&16.36\%\\
\textit{NodeTrans}&32.18&21.50&20.22\%&30.66&19.62&14.80\%&28.54&18.35&12.48\%&41.83&27.02&23.03\%&36.34&23.24&17.13\%&35.22&22.31&16.08\%\\
\hline
\hline
	\end{tabular}}
	\end{table*}

\subsubsection{Parameter Sensitivity}
\begin{figure}[h]
\centering
\subfigure[Number of clusters.]{
\includegraphics[width=0.22\textwidth]{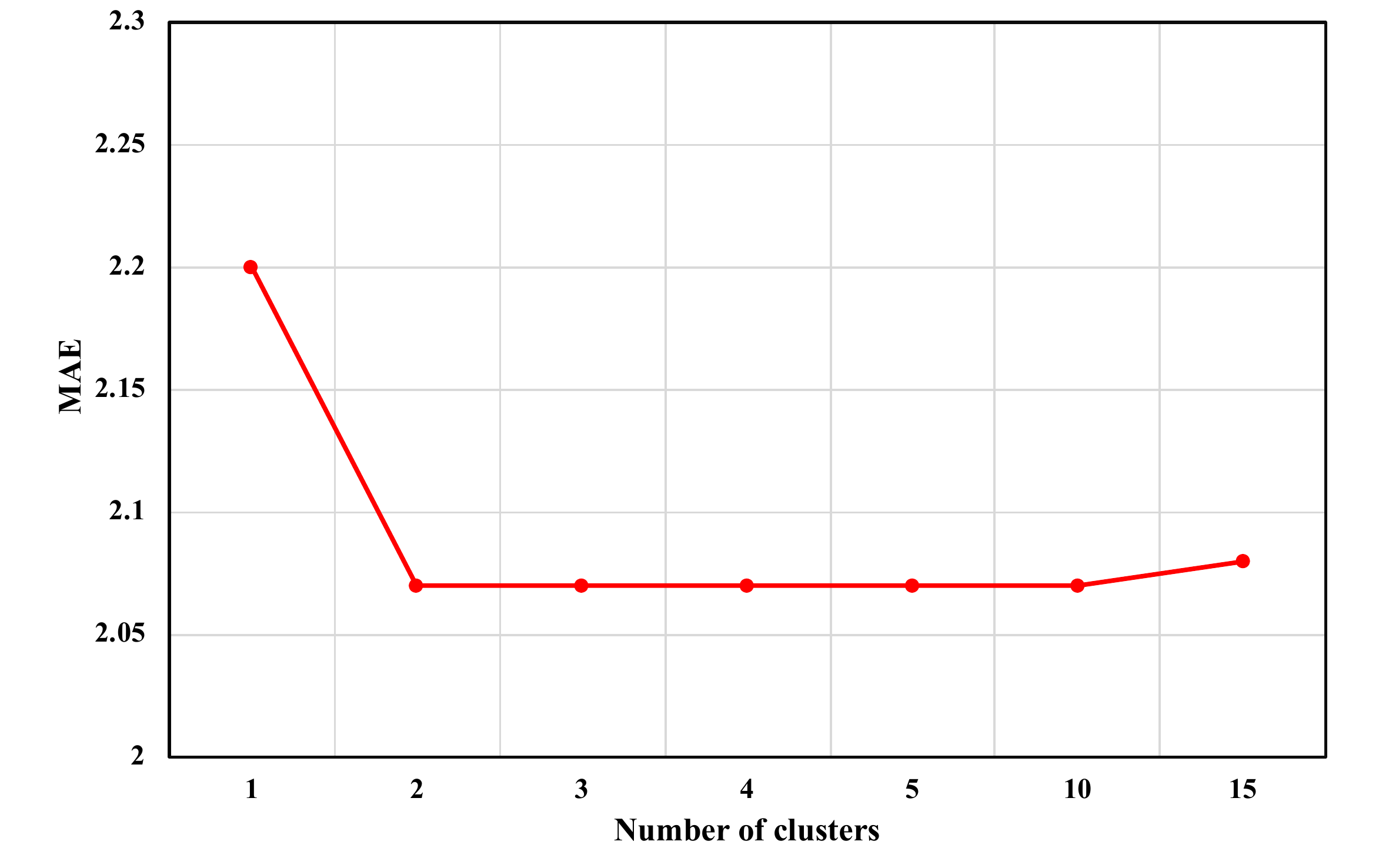}
\label{cluster}
}
\subfigure[Changes of parameter $\alpha$.]{
\includegraphics[width=0.22\textwidth]{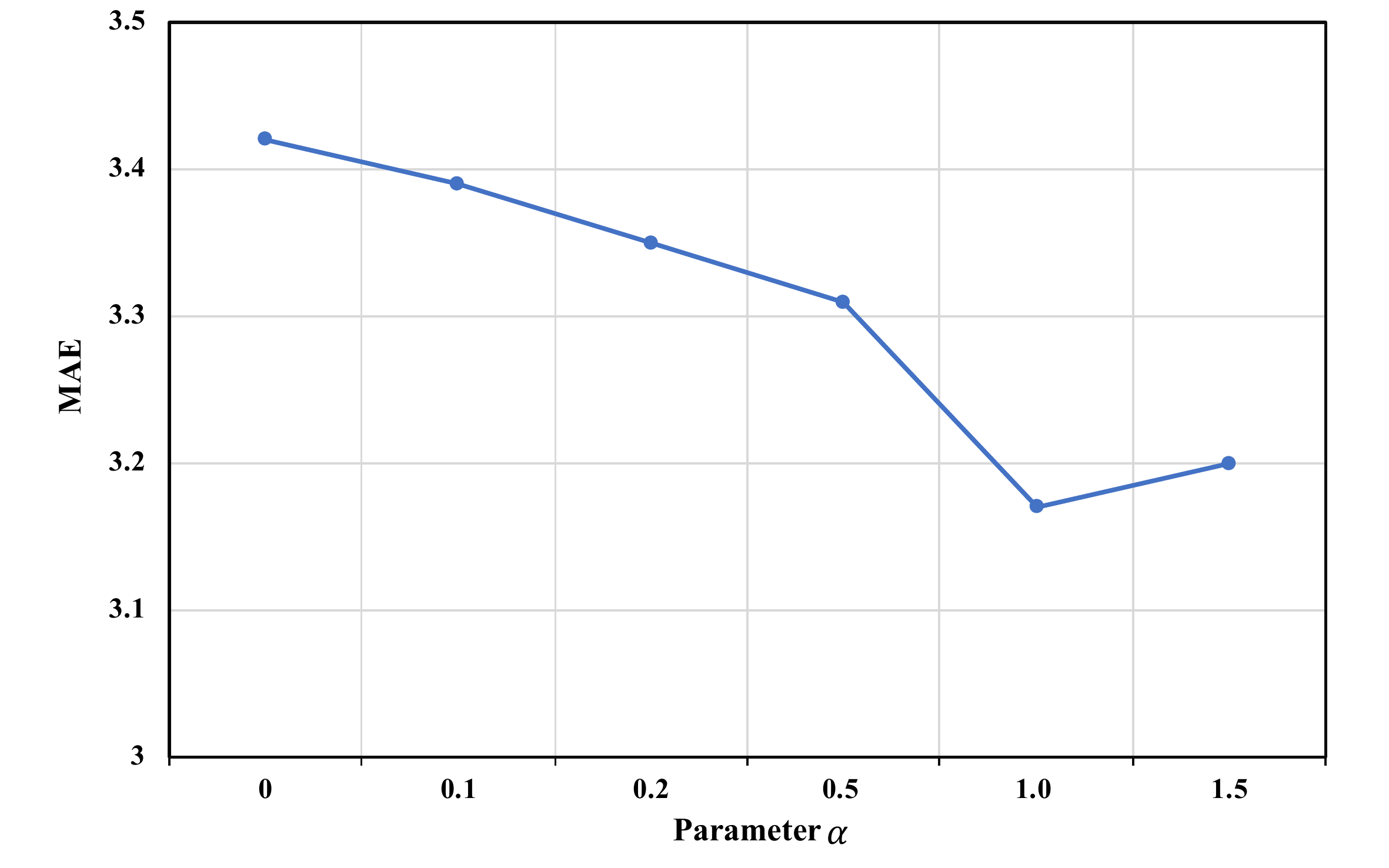}
\label{tradeoff}
}
\caption{Prediction performance of different parameters.}
\label{parameter}
\end{figure}

We use the scenario of 1-day PEMS-BAY to evaluate the effect of our model under two key hyper-parameters. We first evaluate the impact of number of clusters in the pattern-based transfer strategy. As shown in
Fig. \ref{cluster}, the results are sensitive to different cluster numbers, and appropriate values indeed help to
improve the performance of the model.  But note that this parameter depends on particular datasets, and can be used in a data-driven manner to obtain the optimal value. Fig. \ref{tradeoff} shows the effect of trade-off
parameter $\alpha$. It can be seen that our model performs better when the
parameter is set to moderate values. The reason is that small parameter
values will make our approach dominated by the first loss function term in
Eq. (\ref{10}), where the spatio-temporal pattern information is ignored. On
the contrary, if the value of $\alpha$ is set too large, the importance of
spatio-temporal patterns will be over-estimated, and the prediction ability
of the network is neglected.

\begin{figure}[h]
\centering
\subfigure[Cluster 1.]{
\includegraphics[width=0.22\textwidth]{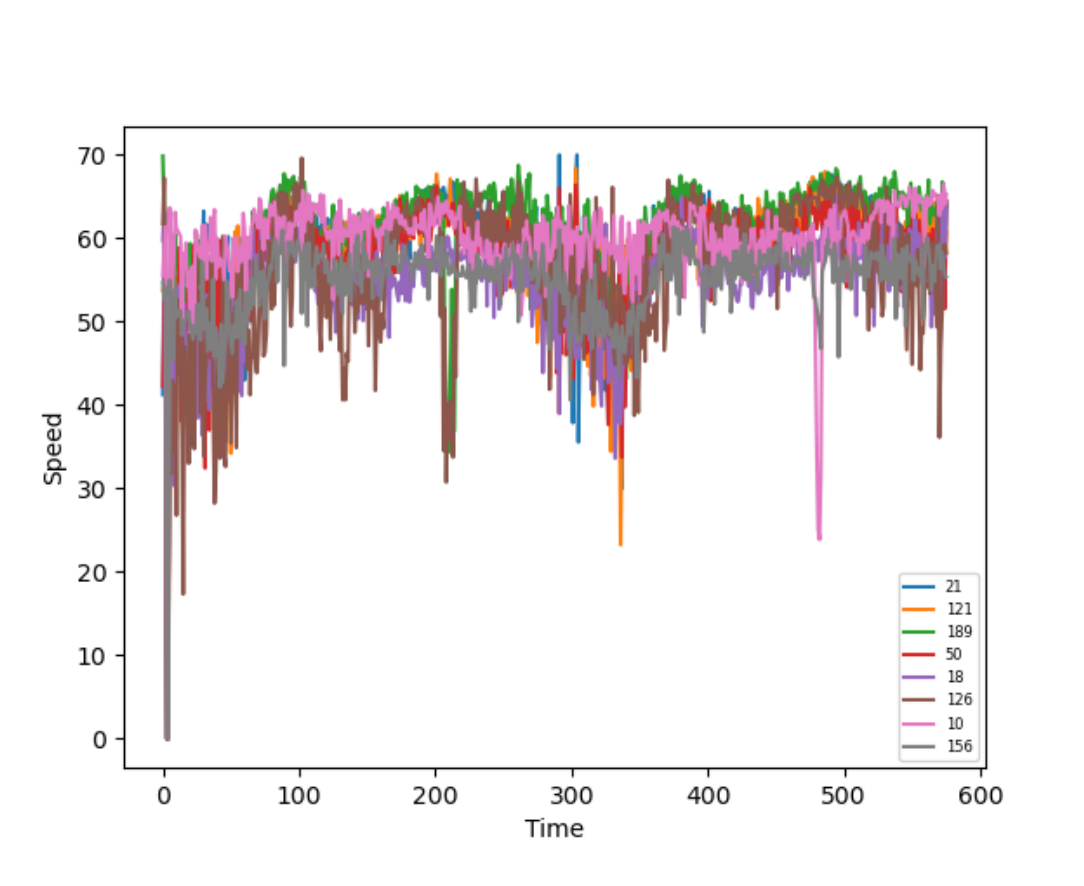}
\label{cluster 1}
}
\subfigure[Cluster 2.]{
\includegraphics[width=0.22\textwidth]{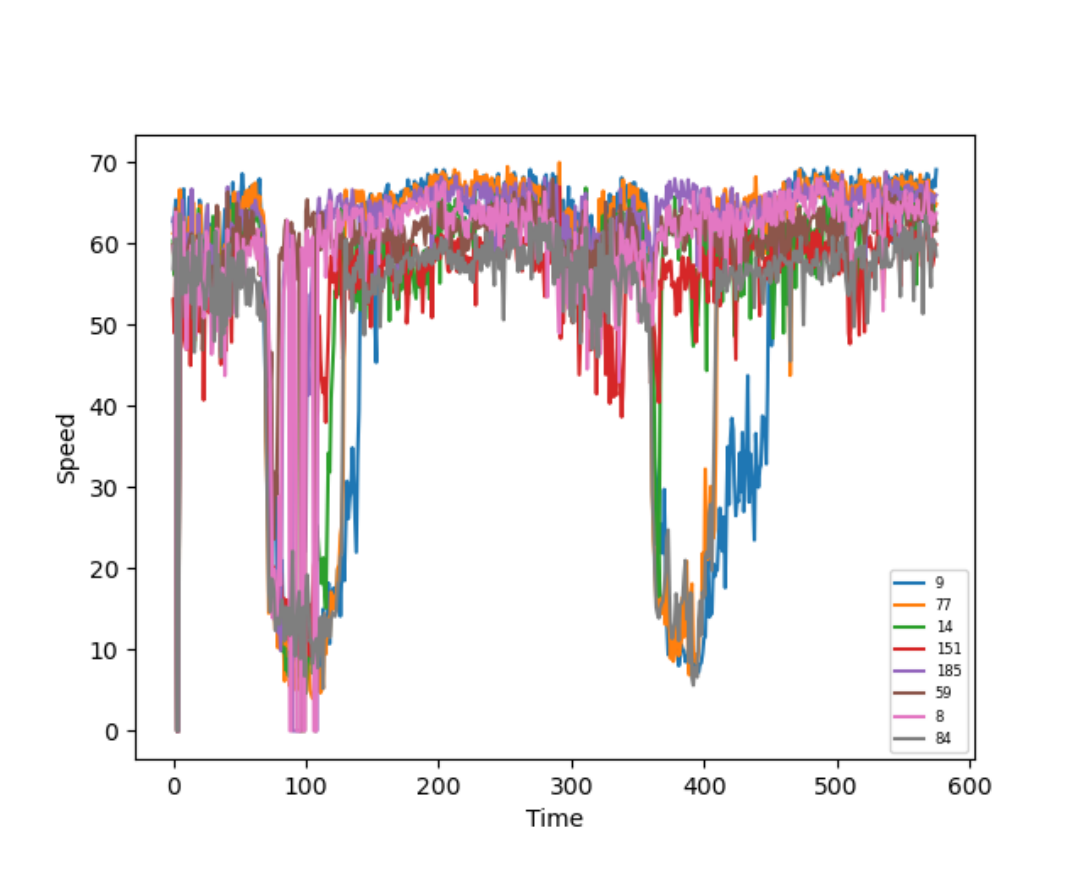}
\label{cluster 2}
}
\subfigure[Cluster 3.]{
\includegraphics[width=0.22\textwidth]{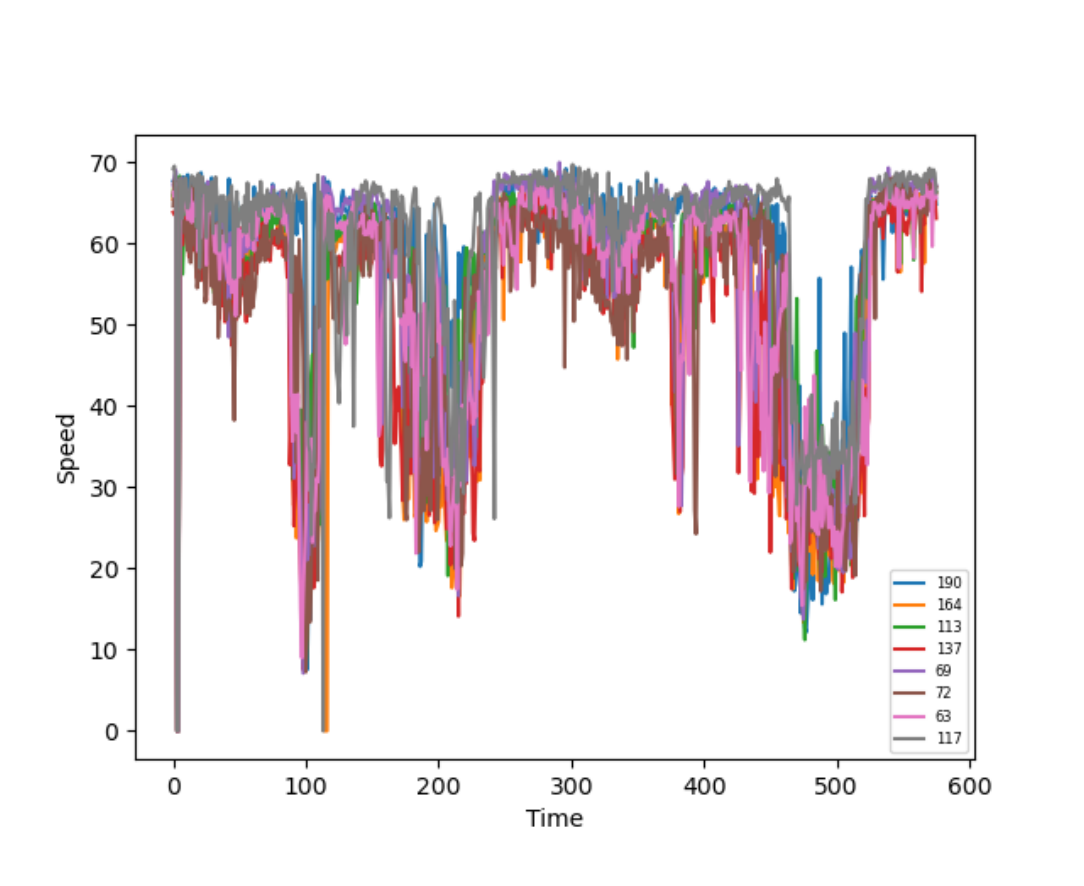}
\label{cluster 3}
}
\subfigure[Cluster 4.]{
\includegraphics[width=0.22\textwidth]{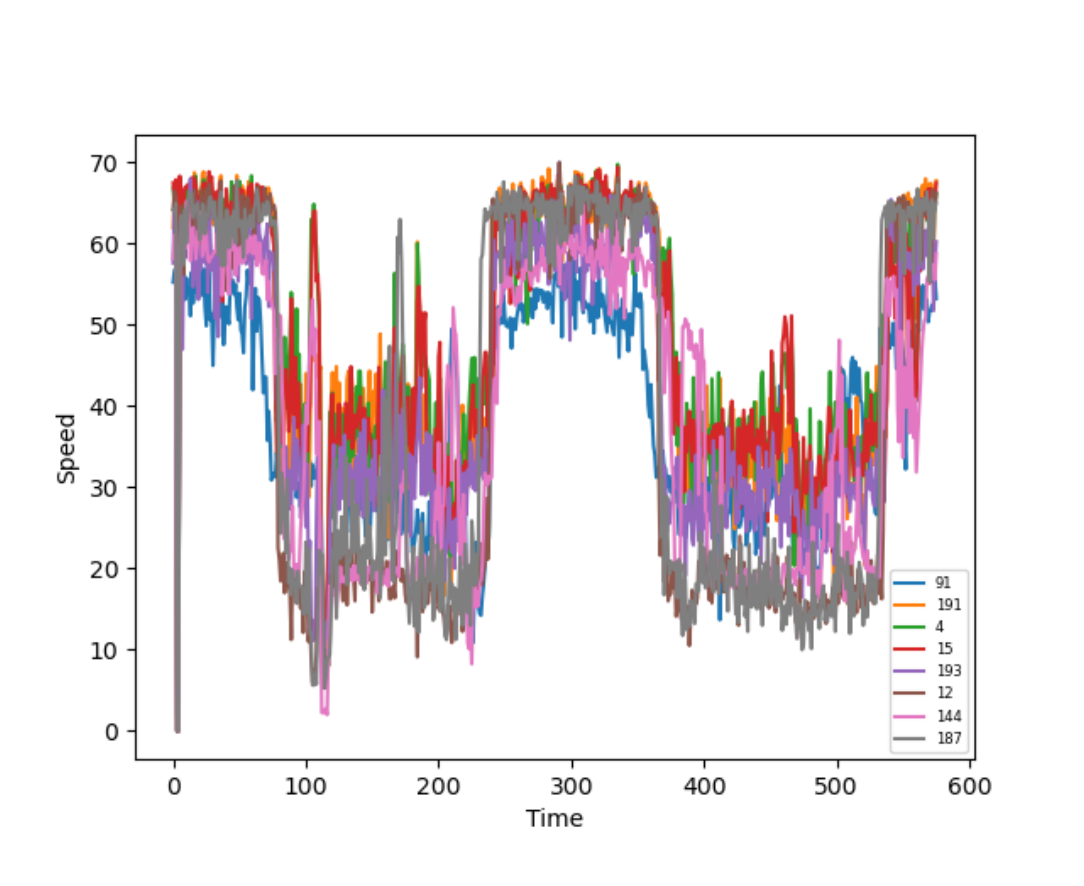}
\label{cluster 4}
}
\subfigure[Cluster 5.]{
\includegraphics[width=0.22\textwidth]{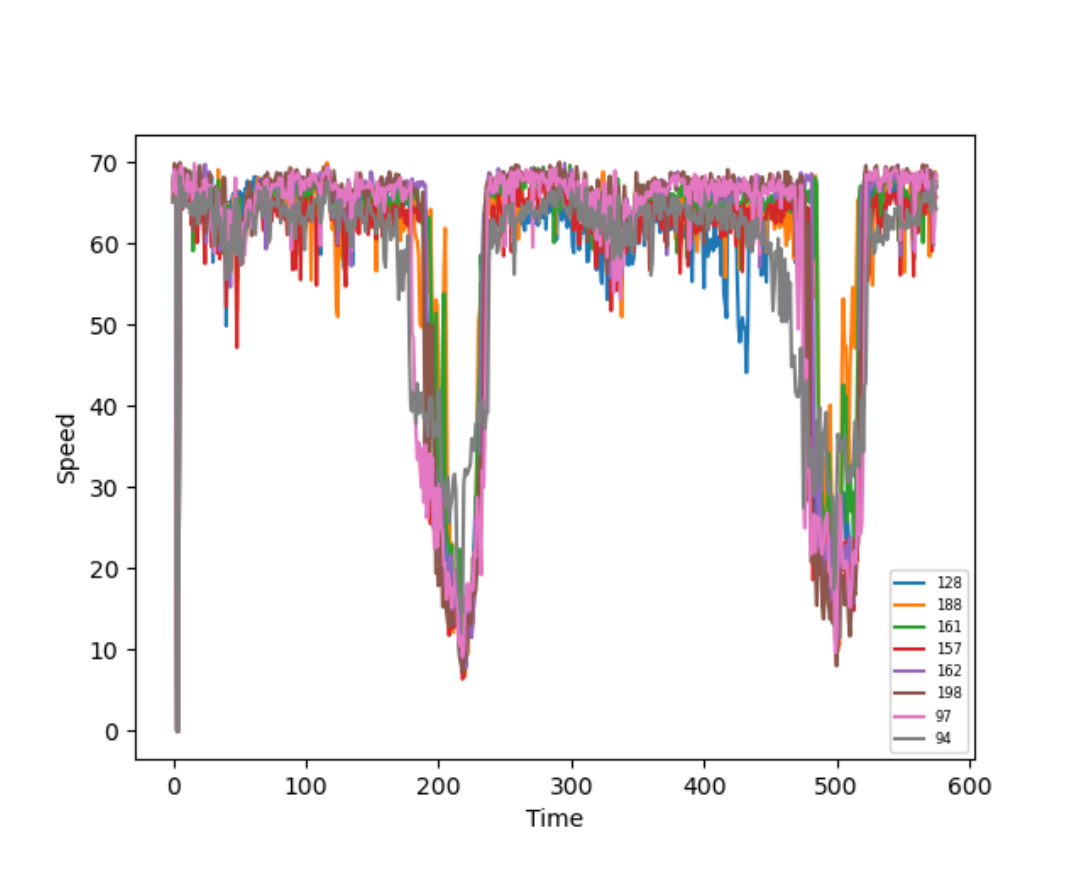}
\label{cluster 5}
}
\caption{Speed changes of different clusters on METR-LA.}
\label{changes}
\end{figure}

\begin{figure}[h]
\centering
\subfigure[Cluster 1.]{
\includegraphics[width=0.22\textwidth]{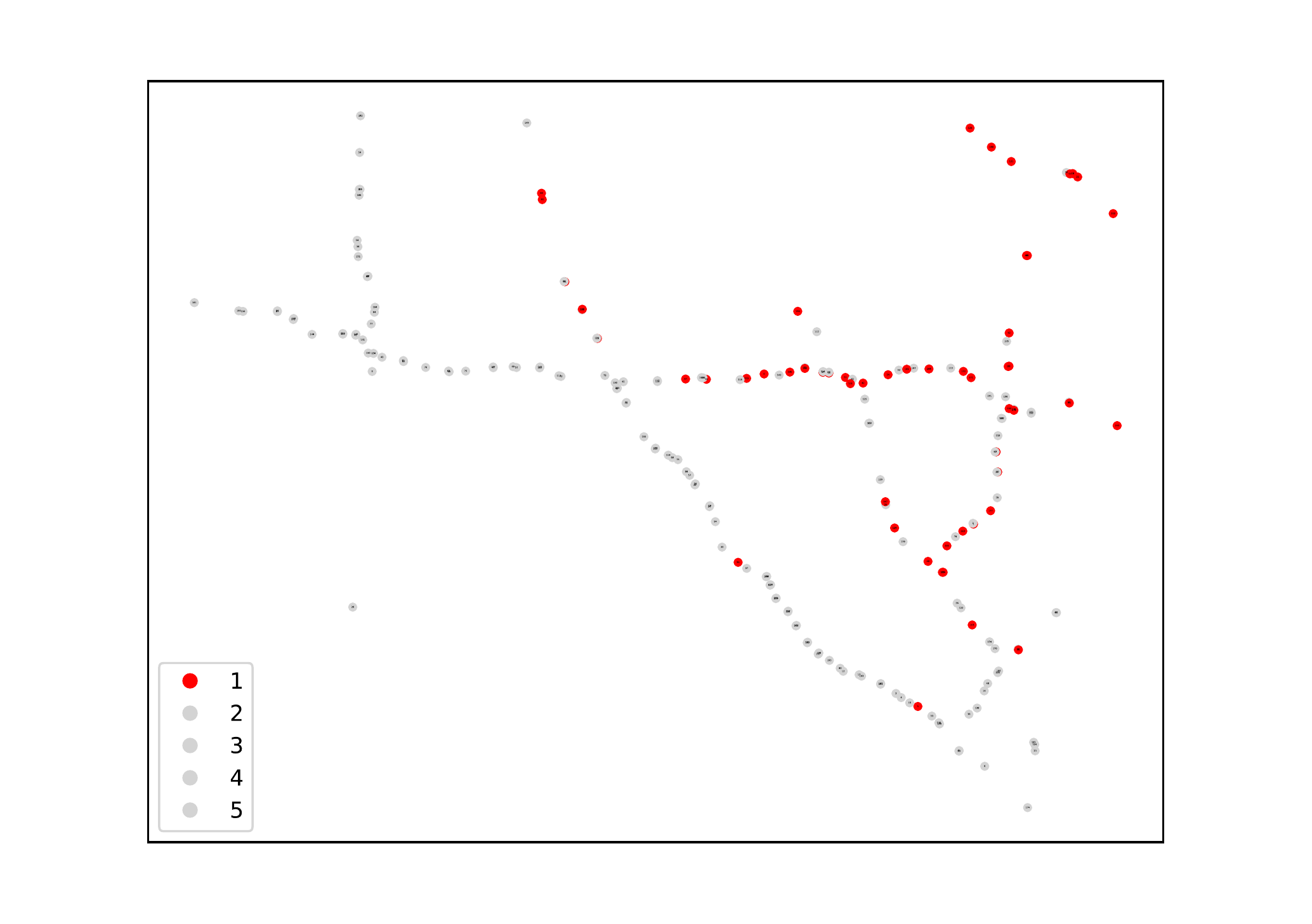}
\label{cluster1}
}
\subfigure[Cluster 2.]{
\includegraphics[width=0.22\textwidth]{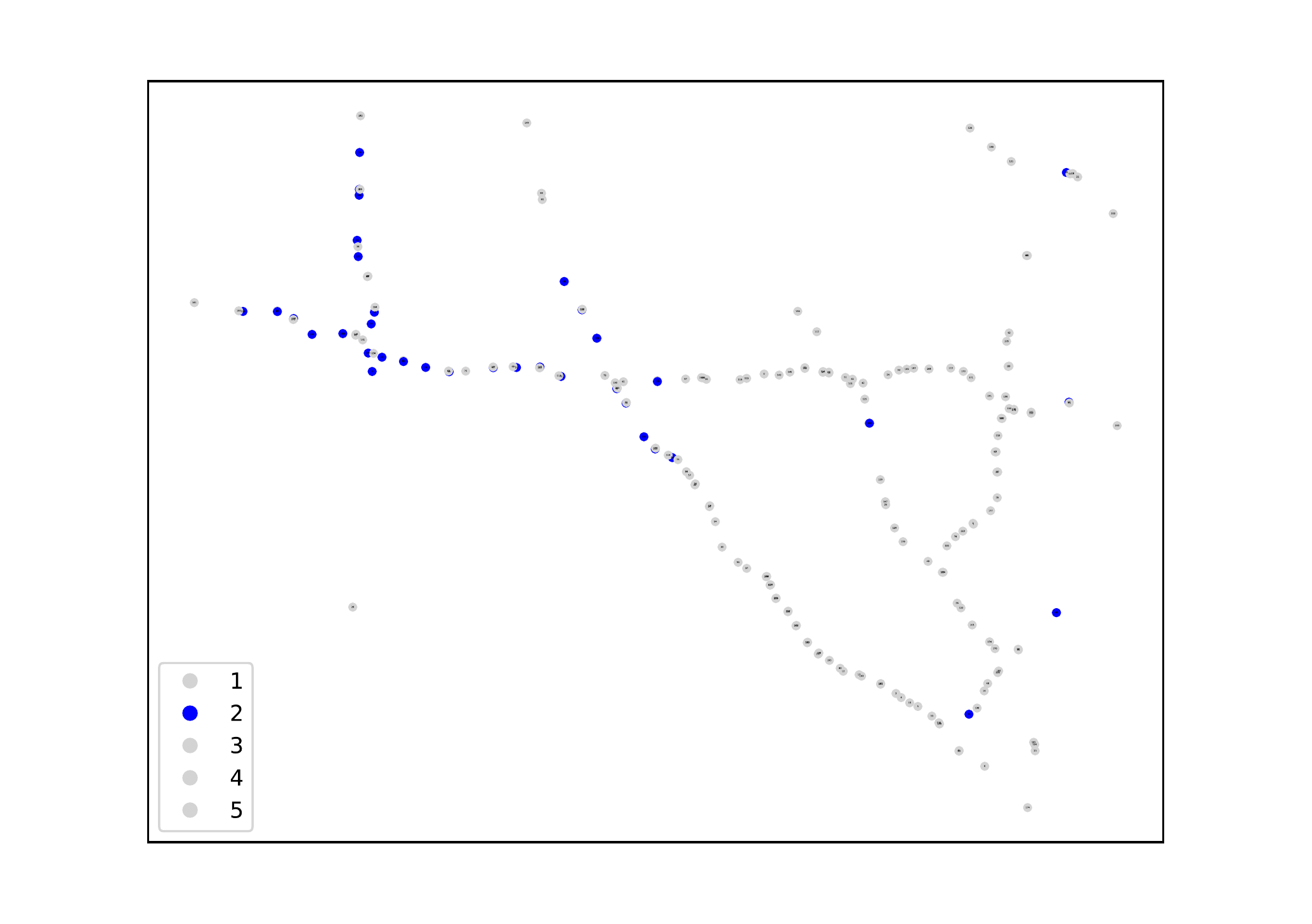}
\label{cluster2}
}
\subfigure[Cluster 3.]{
\includegraphics[width=0.22\textwidth]{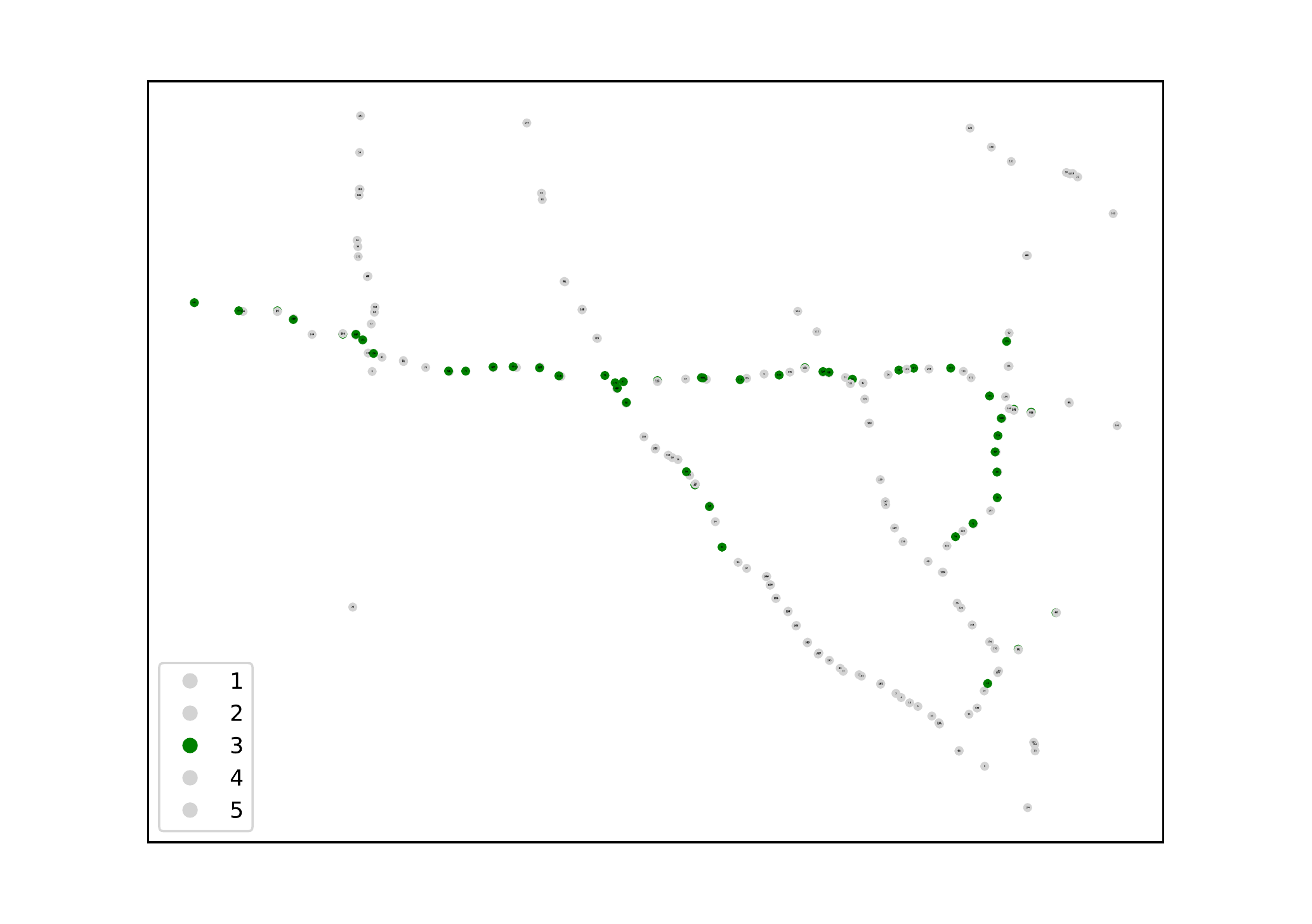}
\label{cluster3}
}
\subfigure[Cluster 4.]{
\includegraphics[width=0.22\textwidth]{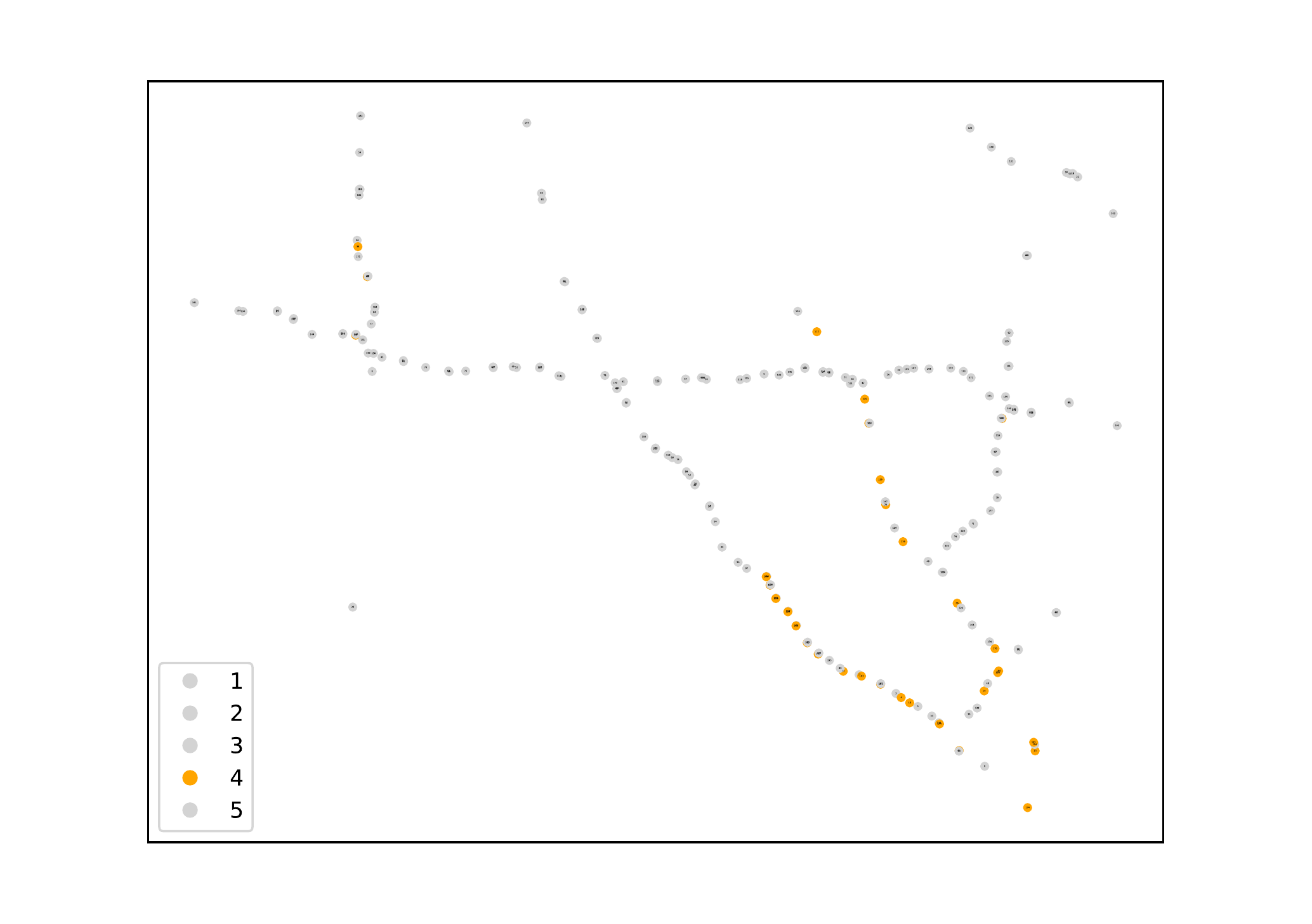}
\label{cluster4}
}
\subfigure[Cluster 5.]{
\includegraphics[width=0.22\textwidth]{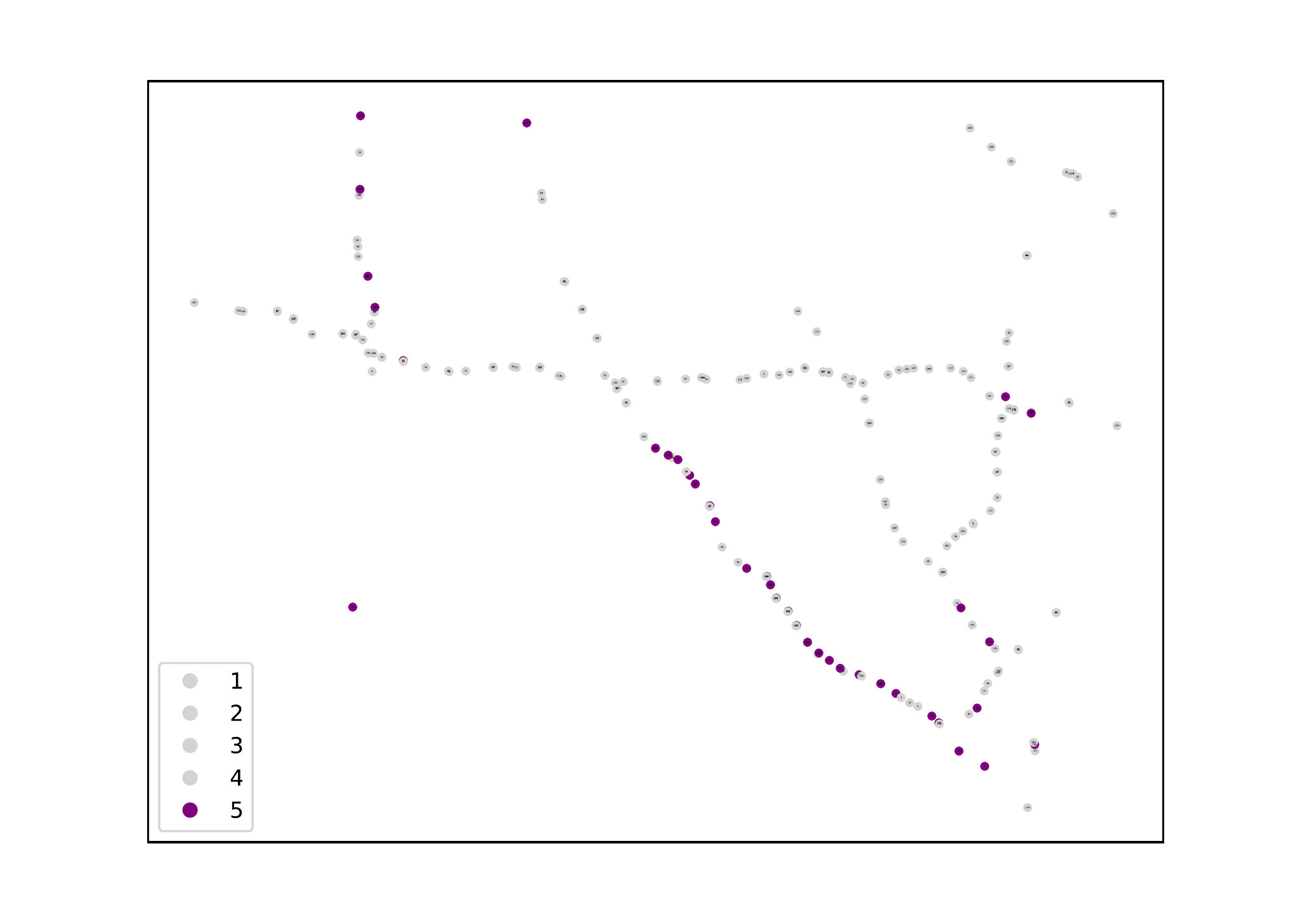}
\label{cluster5}
}
\caption{Geographic visualization of different clusters on METR-LA.}
\label{geo}
\end{figure}

We also visualize the speed changes of different clusters in the source domain 
(e.g. METR-LA) within two days, as shown in Fig. \ref{changes}. 
Several nodes are selected from each cluster to show the trend of changes within two days, and different colors in each subgraph represent different nodes.
It can be observed that the intra-cluster
similarity of traffic condition changes over time is high, and the inter-cluster difference is large. 
In addition, Fig. \ref{geo} visualizes geographic location of each cluster of node on the METR-LA dataset, 
and it can be seen that nodes of the same cluster are relatively
close in geographical location and have similar spatial patterns.

\subsubsection{Computational Cost}
Our proposed spatial-temporal graph neural network (STG-Net) can also bring the computational
gain, greatly reducing the computation time and the number of parameters, as shown in Table \ref{complexity}. It shows the
computational cost of different deep learning models in source domain dataset, where 70\% of the data is used for training, 10\% for validation, and 20\% for testing.
All the
experiments are conducted on the Tesla K80 with 12GB memory, the batch size
of each method is uniformly set to 64, $H$ is set to 12, and we report the
average training time of one epoch. Since STG-Net adopts simple
spatial convolution structure and parallel temporal convolution structure, it can achieve 60\% less than the
second-best approach STGCN.

\begin{table*}[h]
\caption{Computation cost on {PEMS-BAY}.}
\label{complexity}
\centering
\begin{tabular}{p{2.7cm}|p{2.7cm}p{2.7cm}p{2.7cm}}
	\hline
	\multirow{2}{*}{Method} & \multicolumn{2}{c}{Computation time} & \multirow{2}{*}{Number of parameters} \\
\cline{2-3}
	
	                      &Training(s/epoch)&Inference(s)& \\
												\hline
LSTM&18.38&3.57&121157\\
STGCN&35.21&4.66&98592\\
Graph WaveNet&218.84&18.08&247660\\
AGCRN&146.08&13.78&748990\\
HGCN&164.54&18.93&800596\\
STG-Net(our)&18.37&3.63&33210\\
\hline
	\end{tabular}
	\end{table*}

\section{Conclusion}
\label{sec:conc} In this paper, we propose a novel transfer learning
method to address the traffic prediction with few historical data. We
first develop a general spatial-temporal graph neural network and then use
sufficient historical data in the source domain to pre-train the model.
Then, a pattern-based transfer strategy is designed to achieve the
pattern matching between source domain and target domain, improving the
robustness of transfer. Finally, the knowledge learned in the source domain
is used to initialize the model of target domain, and then limited data is
used for fine-tuning under the transfer learning framework.
\textit{NodeTrans} can efficiently solve the problem of data scarcity and
knowledge transfer between graph domains in traffic prediction. Due to its
generality, this paper can be applied to other fields with spatio-temporal
graph data.

\bibliographystyle{IEEEtran}
\bibliography{IEEEabrv,Nodetrans}

\end{document}